\documentclass[letterpaper, 10 pt, conference]{ieeeconf}  

\IEEEoverridecommandlockouts                              

\usepackage{graphics} 
\usepackage{epsfig} 
\usepackage{subfigure}

\usepackage{caption}
\usepackage{float}
\usepackage{multirow}
\usepackage{colortbl,hhline}
\usepackage{amsmath} 
\usepackage{amssymb}  
\usepackage[fleqn,tbtags]{mathtools}
\usepackage{cancel}

\DeclareMathOperator*{\argmax}{argmax}

\usepackage{color,soul}
\usepackage[bordercolor=white,backgroundcolor=gray!30,linecolor=black,colorinlistoftodos]{todonotes}

\usepackage[linesnumbered,ruled,vlined]{algorithm2e}
\usepackage[noend]{algpseudocode}
\usepackage{booktabs}
\usepackage{tabularx}
\usepackage{tabulary}
\usepackage{graphicx} 
\usepackage[export]{adjustbox}

\usepackage[super]{nth}
\usepackage{comment}
\let\labelindent\relax
\usepackage{enumitem}
\usepackage{framed}

\usepackage{makecell}

\title{\LARGE \bf
 Scanning Bot: Efficient Scan Planning using Panoramic Cameras 
 }  

\author{Euijeong Lee$^*$, Kyung Min Han$^*$, and Young J. Kim
\thanks{The authors are with the Department of Computer Science and Engineering at Ewha Womans University in Korea  ${\it \{euijeonglee|hankm|kimy\}@ewha.ac.kr}$. $^*$Equal contributions. \newline \newline
© 2025 IEEE. This work has been accepted to appear in IROS 2025. Personal use of this material is permitted. Permission from IEEE must be obtained for all other uses.}
}

\begin{document}

\maketitle

\begin{abstract}
Panoramic RGB-D cameras are known for their ability to produce high-quality 3D scene reconstructions. However, operating these cameras involves manually selecting viewpoints and physically transporting the camera, making the generation of a 3D model time-consuming and tedious. Additionally, the process can be challenging for novice users due to spatial constraints, such as ensuring sufficient feature overlap between viewpoint frames. To address these challenges, we propose a fully autonomous scan planning system that generates an efficient tour plan for environment scanning, ensuring collision-free navigation and adequate overlap between viewpoints within the plan. Extensive experiments conducted in both synthetic and real-world environments validate our planner's performance against state-of-the-art view planners. 
In particular, our method achieved an average scan coverage of 99\% in the real-world experiment, with our approach being up to 3$\times$ faster than state-of-the-art planners in total scan time.
\end{abstract}

\section{Introduction}

The increasing advancements in mobile robotics research have enhanced robots' ability to improve the efficiency and completeness of outcomes in various active trajectory planning tasks. As a result, modern mobile robots can navigate large-scale environments with minimal human intervention to complete their missions. These applications include autonomous mapping of unknown terrain \cite{CaoZha23}, plant phenotyping \cite{EssBeh23}, 
search and rescue missions \cite{NirNej19}, building information modeling (BIM) \cite{CheChe24} and post-disaster analysis \cite{NagKos21}, among others.  

In particular, efficient view planning based on available information is critical in efficiently achieving reliable results. Indeed, there has been a long history of extensive research regarding how to plan agents' view paths to effectively gather essential information to reconstruct maps or scenes of the environment they navigate.

Although many state-of-the-art view planning systems have been developed over the past few decades, it is difficult to consider view planning a solved problem due to the diversity of applications and problem conditions in the context of mobile robot navigation. Consequently, it remains an active research topic within the robotics community.
For example, in applications such as space exploration or search-and-rescue missions, the quality of the reconstructed model may be less critical than the scanning efficiency. In contrast, for applications such as BIM and digital twins, precise and thorough modeling of the environment is a key component of the problem. In these scenarios, not only is efficient view planning essential, but ensuring sufficient feature overlap between sensor viewpoints is also critical to guarantee a complete reconstruction of the scanned environment.

This paper focuses on the latter challenge, addressing the problem of best-view planning for a panoramic RGB-D camera. Although this camera offers superior reconstruction quality compared to low-end RGB-D cameras, it also introduces critical constraints that must be considered to effectively solve the underlying reconstruction problem, as outlined below.

\begin{enumerate}
    \item The selected viewpoints should ensure complete scene coverage across the entire space.
    \item The number of viewpoints must be minimized, and the planned trajectory must ensure collision-free navigation.
    \item Two neighboring viewpoints in the viewpoint sequence must be within a certain distance to ensure reliable feature matching and construct a high-quality 3D model.
\end{enumerate}

The first and second constraints are common, as similar rules are commonly applied in most view planning systems. 
However, the third constraint is unique to our problem of using a panoramic camera, such as Matterport Pro2, and introduces additional complexity. 
Unfortunately, this constraint hinders us from directly applying previous view planning systems to our problem, as most of these approaches prioritize minimizing trajectory length without accounting for the interactions among viewpoints. 

Furthermore, in the absence of an automated method, a human expert must manually operate the panoramic RGB-D camera, typically selecting viewpoints based on a grid pattern or a similar approach. In addition, the camera must be physically transported to each location and fixed in place, making the process tedious and time-consuming. The challenge becomes even more daunting when large scene models need to be produced to scale up datasets \cite{ChaZha17} for deep learning-related research.

To address these challenges, this research aims to develop a fully autonomous indoor mobile planner, called the scanning bot, that is equipped with a panoramic RGB-D camera. The robot can explore the unknown environment to construct a 2D map using our previous work, Autoexplorer \cite{HanKim22}. Once the map is completed, our novel view planner selects the optimal viewpoints and determines their visiting order. Finally, the scanning bot stitches image data from each viewpoint to reconstruct a 3D model of the observed scenes.
To achieve these goals, we propose a novel view planner with the following contributions:
\begin{itemize}

    \item We propose a novel, fully autonomous mobile scan planning system, \textit{scanning bot}, for reconstructing 3D models of scenes using a panoramic RGB-D camera.  
    
    \item We propose a fast and efficient greedy coverage planner that guarantees comprehensive scanning coverage. Our approach minimizes the number of selected viewpoints while positioning them as closely as possible to ensure sufficient feature overlap among camera views.
    
    \item We propose a new collision-free and visibility-aware path planner that maintains sufficient feature overlap between neighboring viewpoints, ensuring an optimal tour plan for a high-quality reconstruction. 
    
    \item We extensively compared our method with state-of-the-art view planning approaches. This study offers a concise overview of well-established methods and evaluates their performance in the context of the proposed view planning problem. The results demonstrate that our method substantially outperforms existing approaches in terms of efficiency in most scenarios while maintaining over 99\% coverage rate.

\end{itemize} 

\section{Related Work}\label{sec:Related_work}

\subsection{Coverage Path Planning (CPP)}\label{sec:CPP}
CPP focuses on generating an optimal coverage trajectory that is both shortest and collision-free. Given its significance in real-world applications, research in this area remains highly active. It continues to evolve due to its wide range of modern applications, including home cleaning robots, farm-land robots, and mine detection tasks.

Boustrophedon cellular decomposition (BCD) \cite{ChoPig98} is one of the fundamental methods that has served as the foundation for many subsequent CPP studies. As such, one significant challenge in improving trajectory efficiency is minimizing the number of turns when the robot generates a zigzag pattern \cite{VanKol19}. Additionally, differential-wheeled robots often waste considerable time finding paths in cluttered environments or navigating through tight spaces, such as narrow passages. To address these issues, \cite{MutEla23} 
explored CPP with reconfigurable robots, enhancing coverage efficiency in such scenarios.
Moreover, map changes frequently occur during navigation due to human intervention or previously undetected static objects. To handle partially unknown maps, \cite{RamSte24} proposed an efficient replanning method based on an integer programming formulation.
It is important to note that CPP tasks can be completed more efficiently when multiple robots collaborate, as demonstrated in \cite{MitSah24}.


\subsection{Informative Path Planning (IPP)}\label{sec:IPP}
While the primary objective of CPP is to ensure complete coverage and trajectory efficiency, IPP focuses on maximizing information gain or minimizing uncertainty within the exploration area. Consequently, IPP solvers are particularly suited for tasks such as search and rescue missions \cite{LuoZho24}, change detection \cite{BlaSap22}, and large-scale surveillance and monitoring \cite{PopNie20}. Gaussian process regression (GPR) has traditionally been the most widely used approach for IPP problems \cite{XiaWac22}, owing to its probabilistic nature. Research has recently explored deep learning-based IPP solvers, including an attention-based neural network trained using a reinforcement learning (RL) framework \cite{CaoSar23}.

\subsection{Active SLAM}\label{sec:activeSLAM}
Note that an IPP solver can utilize a map configuration as prior knowledge. In such cases, the solver can generate a non-myopic path plan. However, if the exploration space is unknown, the planning must be adaptive, relying on previous observations. IPP solvers designed for unknown space mapping are also known as active SLAM \cite{PacCar23}.
In recent years, numerous new approaches have emerged in active SLAM, particularly driven by advancements in the DARPA Subterranean Challenge \cite{Ack22}.

\subsection{Autonomous Scene Reconstruction}
Scene reconstruction using an RGB-D camera sensor has been a significant research topic in robotics. Among the proposed methods, \cite{xu17} introduced an approach that reconstructs scenes by leveraging time-varying tensor fields. More recently, 3D model reconstruction has regained attention due to the success of 3D Gaussian Splatting (3DGS) \cite{KerDre23}. One notable application of 3DGS is demonstrated by \cite{YuGol24}, who presented an efficient room-scale reconstruction using multiple indoor robots.


\begin{figure*}[ht!]
{\includegraphics[width=\textwidth]{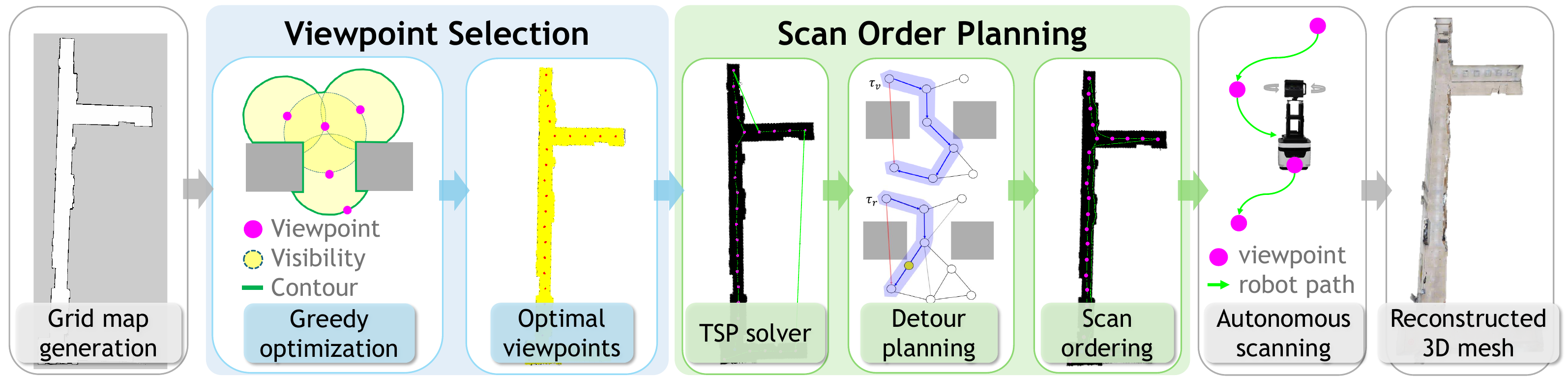}}
\captionof{figure}{ \textbf{Scanning Bot Pipeline.} The set-covering procedure first identifies the optimal viewpoints. This is followed by TSP-based, collision-free view planning to autonomously scan a scene. } \label{fig:pipeline}
\label{fig:pipeline}
\end{figure*}

\section{Problem Formulation}
The main objective of our problem is determining the collision-free optimal view trajectory $\mathcal{T^*}$ that {\em covers} the entire scene. To achieve this, we decompose the problem into two phases: (1) optimal viewpoint $\mathcal{V^*}$ selection and (2) generating $\mathcal{T^*}$ from $\mathcal{V}^*$. 

\vspace*{-0.5em}
\subsection{Optimal Viewpoint Selection}\label{subsec:setcovering_formulation}
A 2D grid map $\mathcal{M}$ consisting of two types of grid cell classes: (i) occupied cells $\mathcal{O}$ corresponding to obstacles, (ii) unoccupied or free cells $\mathcal{F}$ corresponding to the collision-free space.
Such a map can be given or autonomously constructed \cite{HanKim22}.

$x_i$ is said to be \textit{visible} to $x_j$, $x_i \mapsto x_j$ (or $x_j \mapsto x_i$) if
\begin{subequations}
\begin{align}\label{eq:visibility_cond}
\vspace*{-1.0em}
\mathbf{\Lambda}=tx_i+(1-t)x_j \in \mathcal{F}, \; 0\le \forall t \le 1 \; \mathrm{and} \; ||\mathbf{\Lambda}|| \le r,  
\end{align}
\end{subequations}
\noindent where $r$ is a sensor-range constant.
A cell $x_i \in \mathcal{F}$ {\em sees} a set of free cells $\mathcal{S}_i \in \mathcal{F}$, denoted as $x_i \mapsto S_i$, when  $\forall x \in \mathcal{S}_i, x_i \mapsto x$.
Our first goal is to find a minimal set of "viewpoint" cells, $\mathcal{V}^*=\{x_i\in \mathcal{F}\}$ such that $\mathcal{F} \subset \underset{x_i \mapsto \mathcal{S}_i}\bigcup \mathcal{S}_{i}$. 
This optimization problem can be formulated as a set covering problem \cite{Ata10}:
\begin{subequations}
\label{eq:vertex_cover}
\begin{align}
\min_{} \quad & \sum\limits_{\forall i}{\alpha_i\cdot \bar{x}_i}, \quad \bar{x}_i \in \{0,1\}, \alpha_i \in \mathbb{R}^+  \label{eq:objfunc} \\
\textrm{s.t.} \quad & \sum\limits_{i}^{}{\bar{x}_i} \geq 1, \quad 1\le \forall i \le \mathcal{|F|}  \label{eq:setcover_cond1} \\
& \bar{x}_i - \sum\limits_{j \neq i}{ \bar{x}_j } \leq 0 \quad \text{for} \: \rho( x_{i}, x_{j} ) \le r, \: \forall i, \label{eq:setcover_cond2} 
\end{align}
\end{subequations}
where $\bar{x}_i$ corresponds to a viewpoint $x_i \in \mathcal{F}$, $\alpha_i$ refers to the significance of $x_i$, and $\rho$ is a distance function between two viewpoints, $x_i \mapsto x_j$. 
Eq. \ref{eq:objfunc} and \ref{eq:setcover_cond1} form a typical set covering problem, but Eq. \ref{eq:setcover_cond2} 
complicates the combinatorial optimization problem with geometric constraints. Finally, $\mathcal{V}^*=\{\forall x_i | \bar{x}_i \neq 0 \}$ from the solution of Eq. \ref{eq:objfunc}.

\subsection{Optimal Scan Ordering}\label{subsec:tour_planning_formulation} 
This subsection formulates our second goal: to generate an optimal scan tour plan from $\mathcal{V}^*$.
Formally, our tour optimization problem is:

\begin{subequations}
\label{eq:tsp}
\begin{align}
\displaystyle \min
\quad&{\sum\limits_{i}\sum\limits_{j\neq i} \beta_{ij} \bar{e}_{ij}} \label{eq:tsp_obj}\\
\textrm{s.t.}
\quad&\displaystyle \sum\limits_{j \neq i}^{n}{ \bar{e}_{ij}} = 1 \:\: \textrm{and} \:\: \sum\limits_{i \neq j}^{n}{ \bar{e}_{ij}} = 1  \label{eq:tsp_cond1}\\
\quad& \forall i, \; \exists j < i, \; \rho(x_i, x_j) \le r \label{eq:tsp_cond2}
\end{align}
\end{subequations}
Here,  $\bar{e}_{ij} \in \{0,1\}$ represents a tour sequence from $x_i \in \mathcal{V}^*$ to $x_j \in \mathcal{V}^*$, and
$\beta_{ij}$ represents the cost for choosing $e_{ij}$. For instance, $\beta_{ij} = \infty$ if $x_i$ is not visible from $x_j$. We call such an edge $e_{ij}$ with $\beta_{ij}=\infty$ {\em infeasible}.
Eq. \ref{eq:tsp_obj} and \ref{eq:tsp_cond1} constitute a typical TSP problem, respectively, but our problem requires the additional precedence constraint (Eq. \ref{eq:tsp_cond2}) where each viewpoint in the final tour plan must have a visible viewpoint among its predecessors. A solution to Eq. \ref{eq:tsp_obj} is the final tour $\mathcal{T}^*=\{\bar{e}_{12}, \cdots, \bar{e}_{n-1,n}\}, \forall \bar{e}_{i, i+1} \neq 0$.

\section{Autonomous View Planning}
\subsection{Overview} \label{subsec:overview}
Fig. \ref{fig:pipeline} shows the full pipeline of our method consisting of viewpoint selection and scan order planning to be described in the following Secs. \ref{subsec:our_viewpoint_collection} and \ref{subsec:our_tourplan}, respectively. 

\subsection{Viewpoint Selection} \label{subsec:our_viewpoint_collection}
\begin{figure}[htb!]
\begin{centering}
\includegraphics[width=0.65\columnwidth]{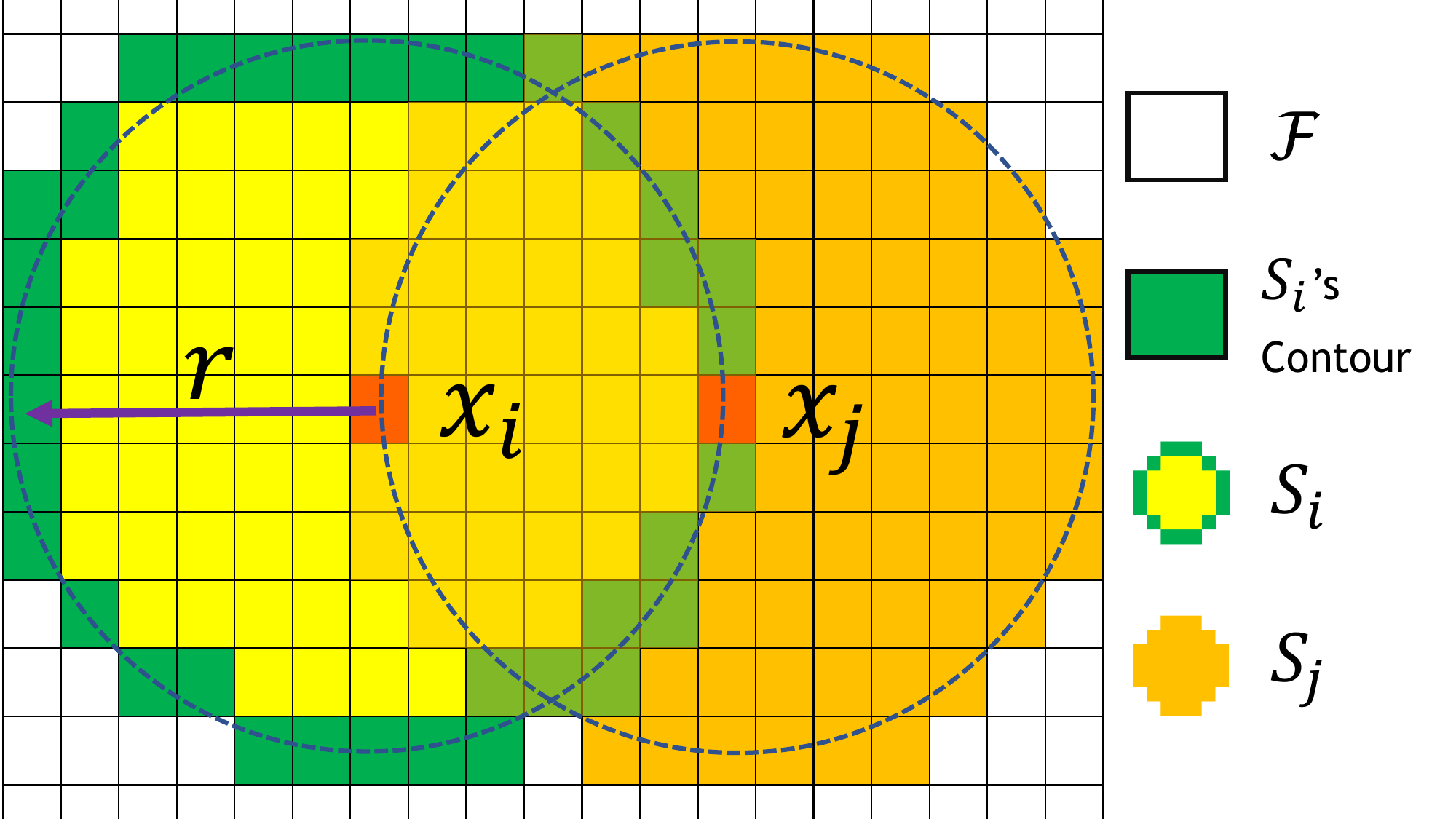}
\caption{ \textbf{Greedy set-covering.} $x_i$ covers the map with $\mathcal{S}_i$. The green cells represent $\mathcal{S}_i$'s contour cells. The next viewpoint $x_j$ is selected from these contour cells based on $\phi(\cdot)$.}   \label{fig:vizsetcover}
\end{centering}
\vspace*{-1.0em}
\end{figure}

Due to the additional complexity introduced by Eq. \ref{eq:setcover_cond2}, the standard set-cover formulation is unsuitable for our problem. Furthermore, the size of $\mathcal{F}$ in our problem also hinders efficient optimization.
To address these challenges, we propose a greedy set covering-based approach to compute $\mathcal{V}^*$ as explained below
(also illustrated in Fig. \ref{fig:vizsetcover}):

\begin{enumerate}
\item Initially, using \cite{EppKon10}, we compute a universe of free cell sets $\mathcal{U} = \{\mathcal{S}_i | \forall x_i \in \mathcal{F}, x_i \mapsto \mathcal{S}_i  \}$. 
Pick $x_0$ where $\mathcal{S}_0$ is the largest in $\mathcal{U}$, and set $\mathcal{V}^* = \{x_0\}$.

\item Determine 
the contour cells $\partial \mathcal{S}_i$ of $\mathcal{S}_i$
utilizing DFFP \cite{HanKim24}. 

\item For each $x$ in $\partial \mathcal{S}_i$, evaluate $\phi(\cdot)$, which depends on the size of the set and the distance to the nearby obstacles:
\begin{equation}
\label{eq:viewpoint_eval}
\phi(x) = |\mathcal{S}_i| / \max_{i}|\mathcal{S}_i| - {e^{-\lVert x - \mathbf{o} \rVert _2} },
\end{equation}
where $\mathbf{o}$ represent the position of the obstacle cell closest to $x$. 
Choose $x_j = \argmax_x \phi(x)$ as the next best viewpoint $x_j$ and add $x_j$ to  $\mathcal{V}^*$.

\item We repeat 2)$\sim$4) until $\mathcal{F} \subset \{S_i | \forall x_i \in \mathcal{V}^* \}$.
\end{enumerate}

\subsection{Scan Order Planning }\label{subsec:our_tourplan}


    
        

\begin{figure}[htb!]
\begin{centering}
\subfigure[Visibility graph $\mathcal{G}_v$]
{\label{fig:viz_graph}\includegraphics[width=0.3\columnwidth]{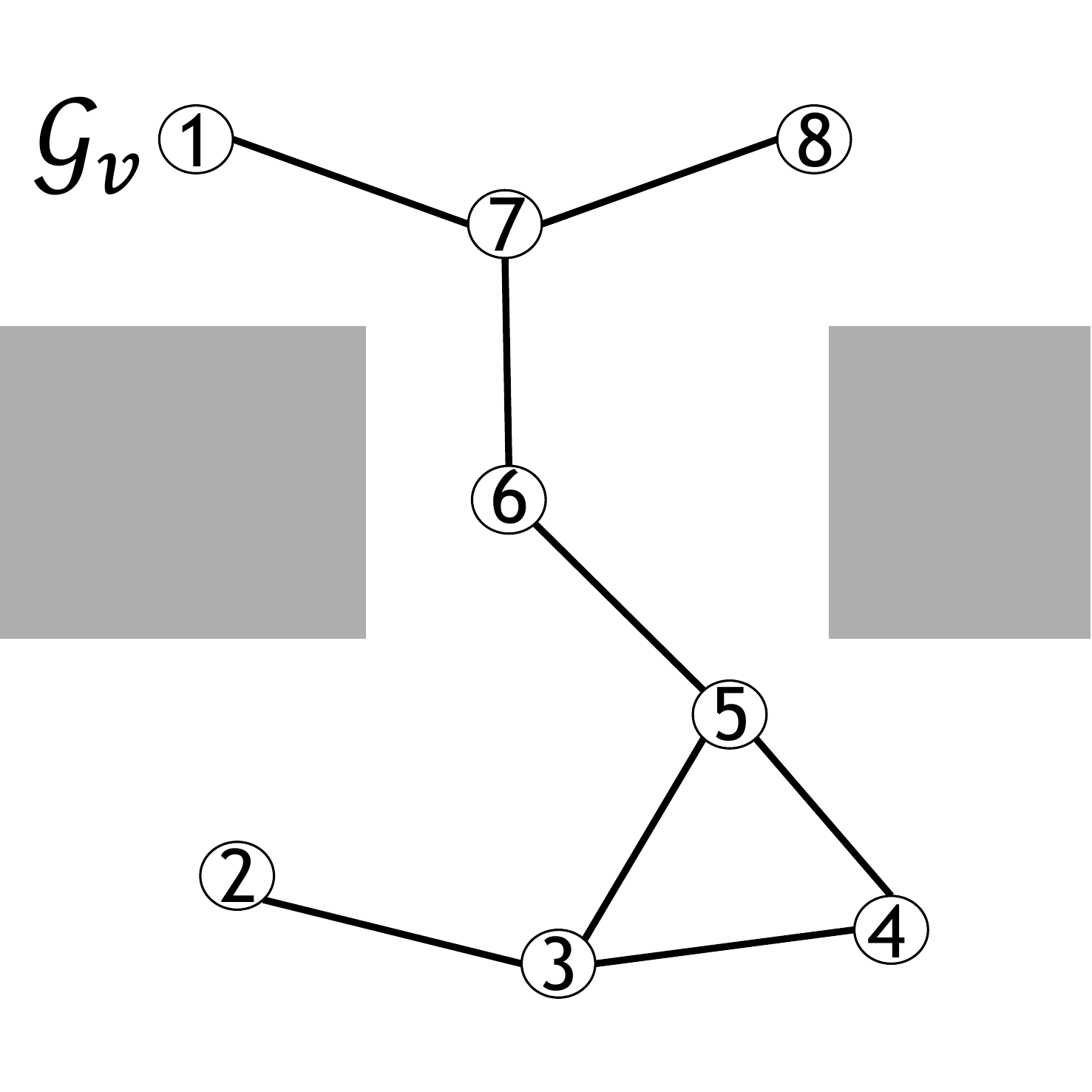}}
\subfigure[$\mathcal{T}$ including an infeasilbe path $\tilde{e}_{12}$]
{\label{fig:infeasible_tsp}\includegraphics[width=0.3\columnwidth]{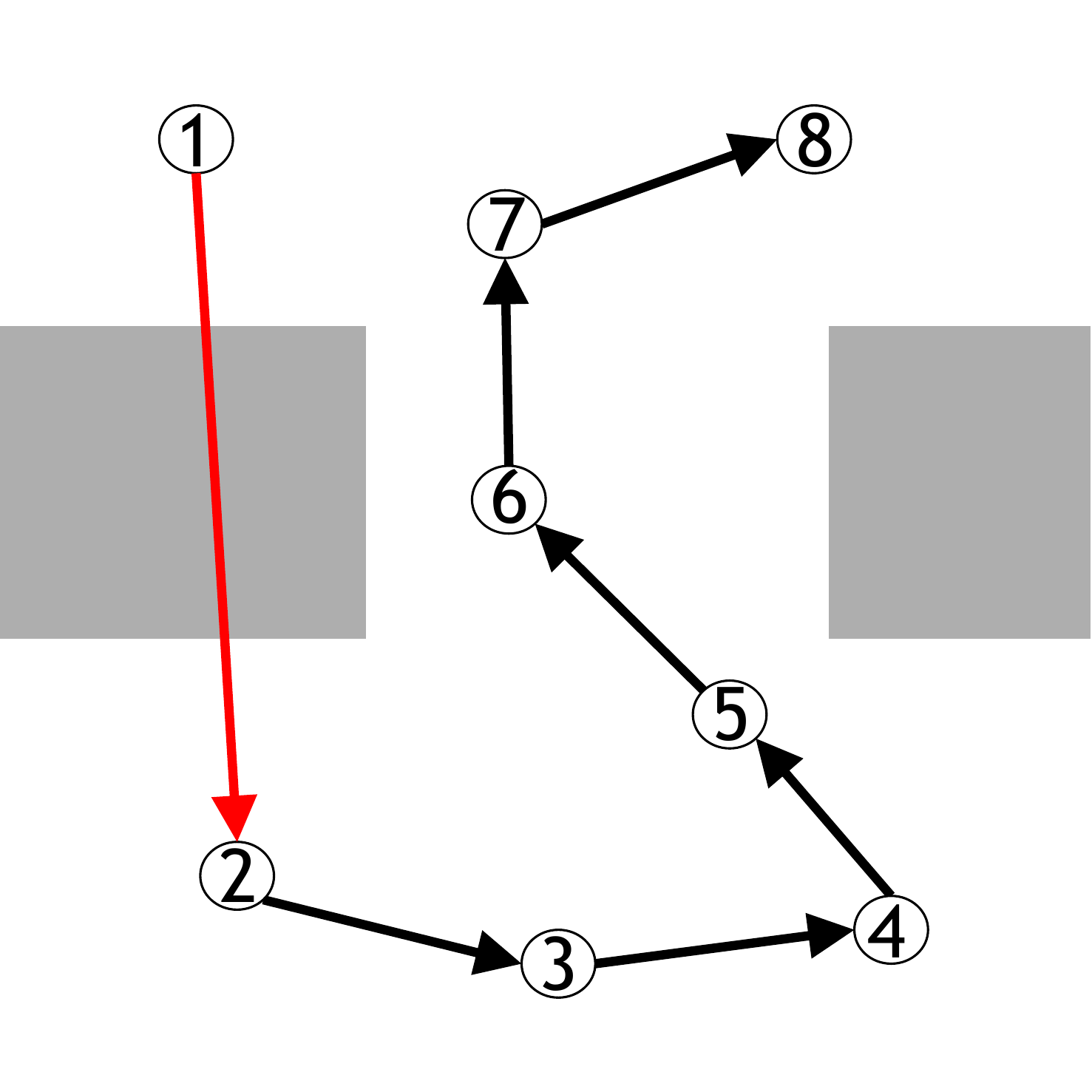}}
\subfigure[$\mathcal{\tau}_v$ (blue detour)]
{\label{fig:tv}\includegraphics[width=0.3\columnwidth]{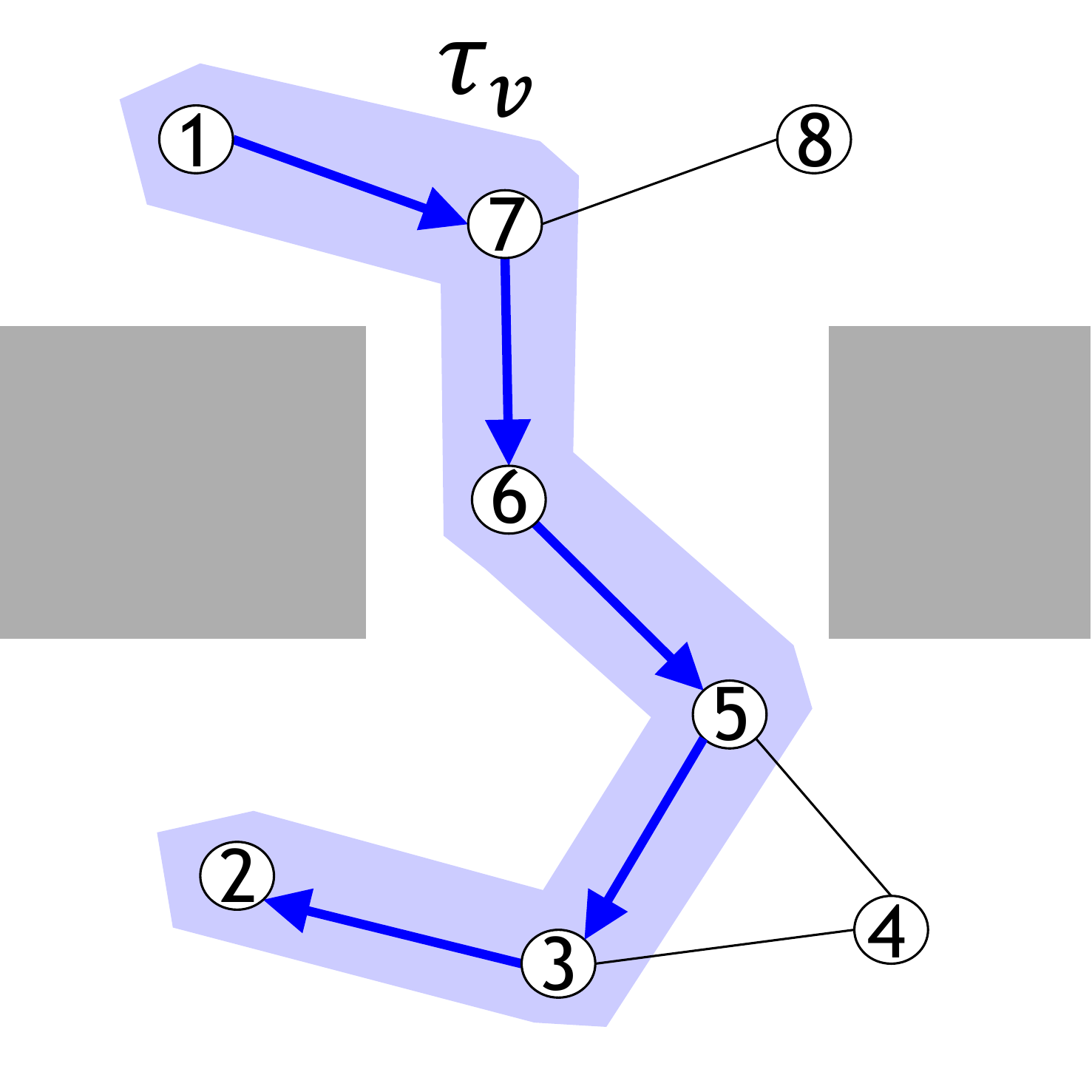}}

\subfigure[Roadmap $\mathcal{G}_r$]
{\label{fig:roadmap_graph}\includegraphics[width=0.3\columnwidth]{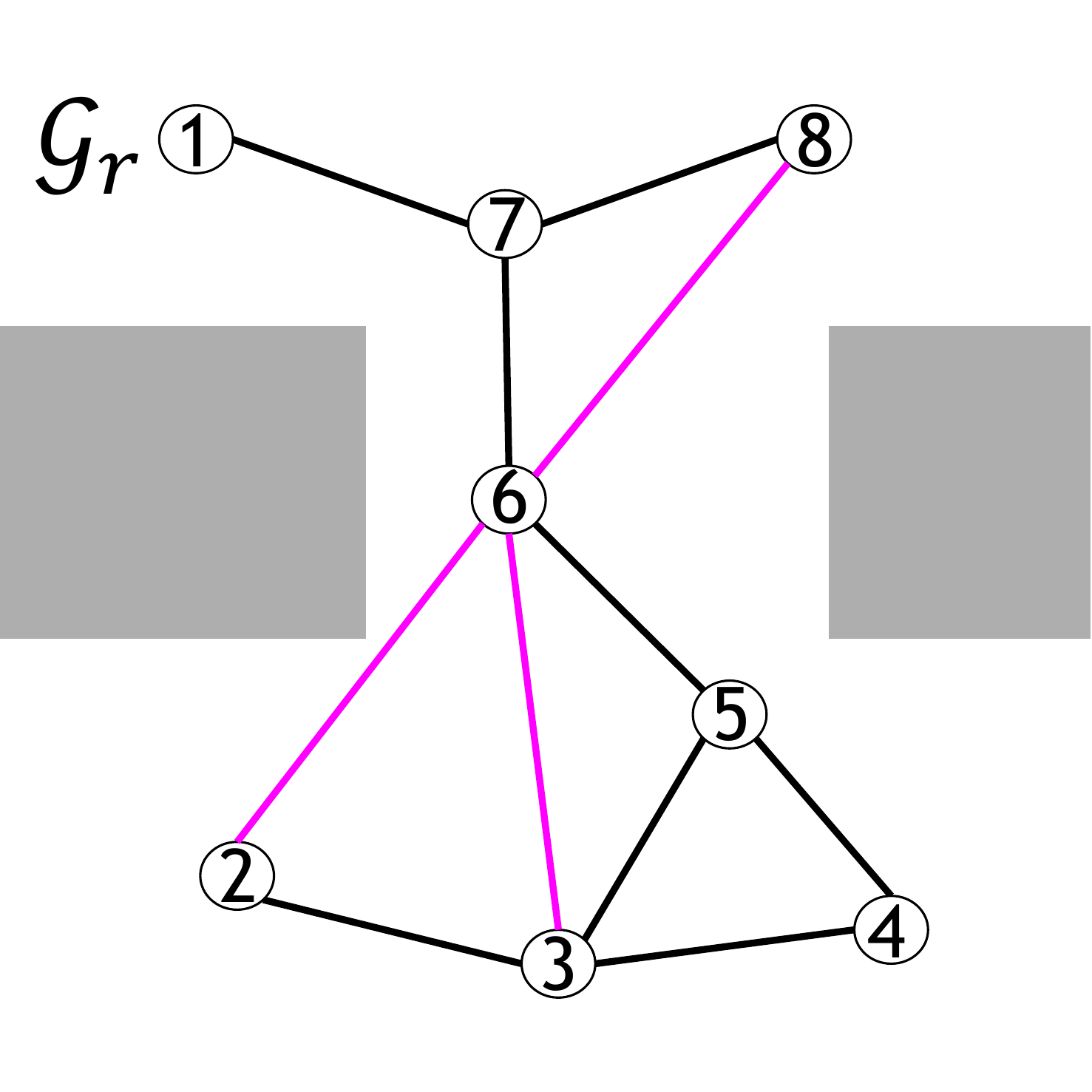}} 
\subfigure[$\mathcal{\tau}_r$ (blue) with the Steiner node ($x_9$)]
{\label{fig:tr}\includegraphics[width=0.3\columnwidth]{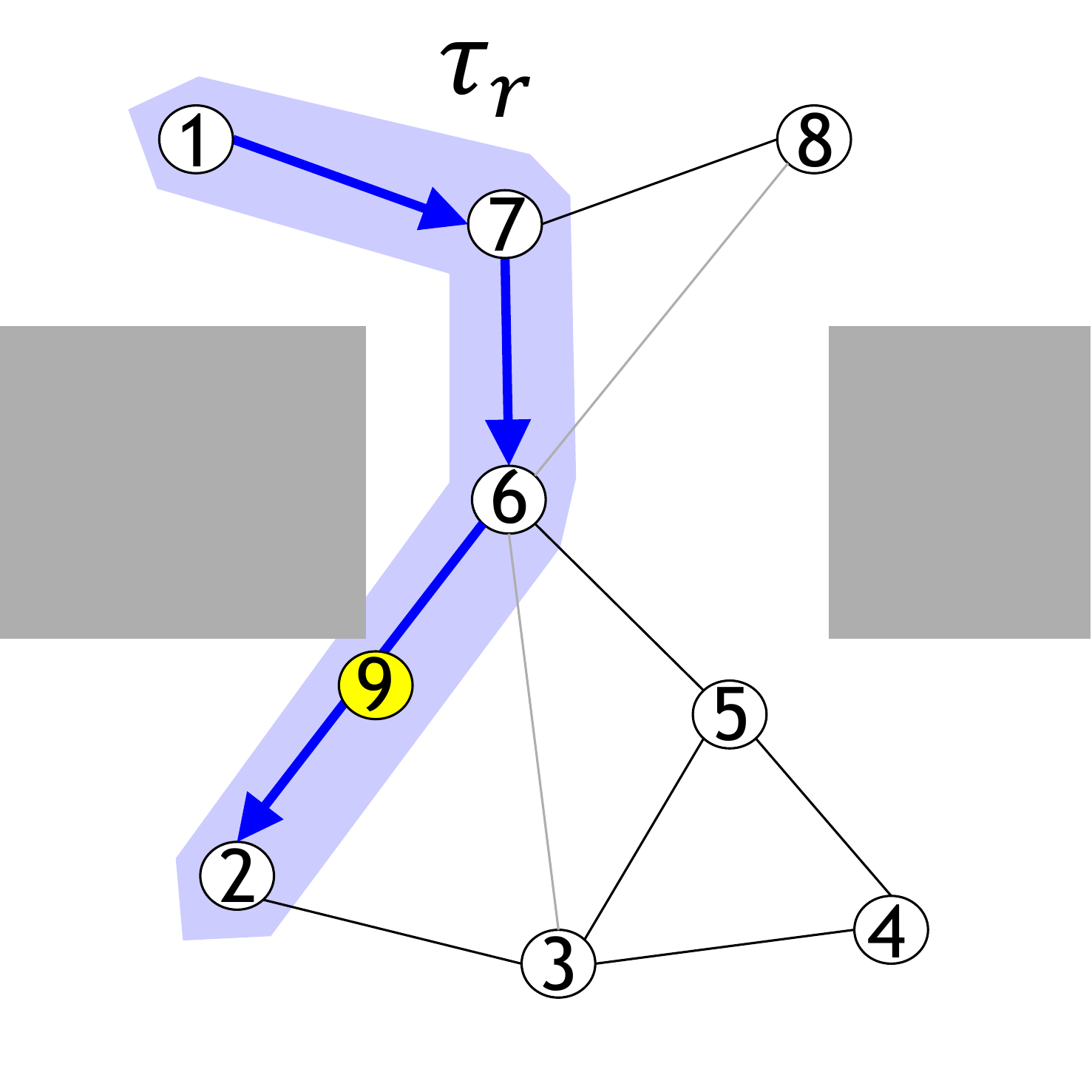}}
\subfigure[The final $\mathcal{T}^*$]
{\label{fig:traj_star}\includegraphics[width=0.3\columnwidth]{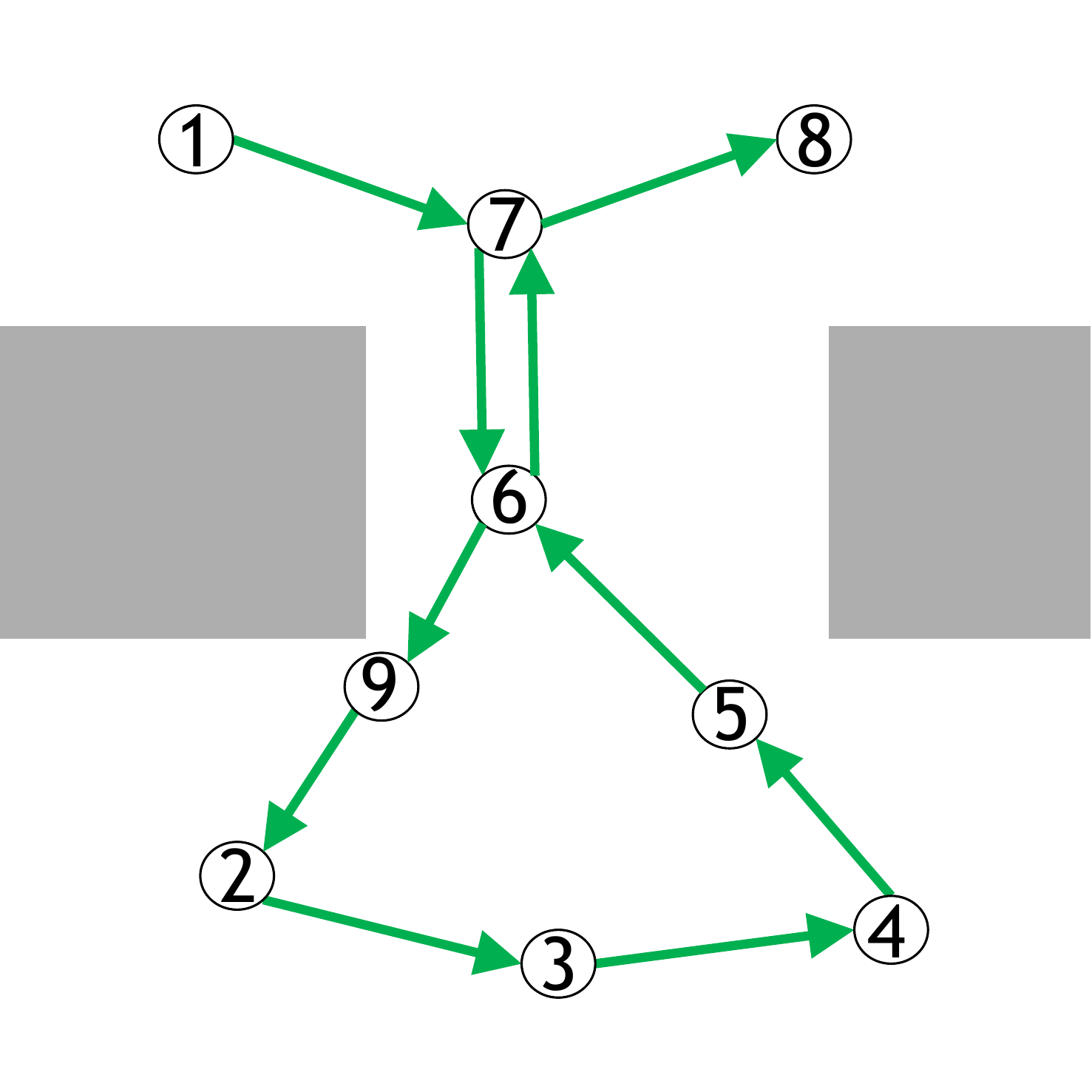}} 

\caption{ \textbf{Detour plan using the roadmap.} (a) $\mathcal{G}_v$ (b) a TSP plan $\mathcal{T}$ with an infeasible edge between $x_1$ and $x_2$. (c) presents the  $\mathcal{\tau}_v$ found in $\mathcal{G}_v$. (d) and (e) show the roadmap $\mathcal{G}_r$ and $\mathcal{\tau}_r$ found on $\mathcal{G}_r$, respectively.  $\mathcal{\tau}_r$ has a new Steiner node $x_9$. (f) the final path $\mathcal{T}^*$.} \label{fig:shortest-path}
\end{centering}
\vspace*{-2.0em}
\end{figure}

We use a TSP solver such as \cite{TSP06} to initialize the scan planning $\mathcal{T}$ using $\mathcal{V}^*$ as input. However, $\mathcal{T}$ may not satisfy the constraints in Eq.~\ref{eq:tsp_cond2}, yielding infeasible paths.
To address this issue, our approach searches for a detour plan $\mathcal{\tau}$ which replaces any infeasible edge $\tilde{e} \in \mathcal{T}$. 
For instance, Fig. \ref{fig:viz_graph} and \ref{fig:infeasible_tsp} show an instance of $\mathcal{G}_v$ and the initial planning $\mathcal{T}=\{e_{12}, \cdots, e_{78}\}$ obtained from the TSP solver starting from $x_1$, respectively. The edge $\tilde{e}_{12}$ in  $\mathcal{T}$ is infeasible, and needs to be detoured.


One approach is to search for a detouring path $\mathcal{\tau}_v$ in the visibility graph $\mathcal{G}_v=(\mathcal{V}^*, \mathcal{E}_v)$, where $\mathcal{E}_v = \{ e_{ij} | \beta_{ij} \neq \infty, \forall ij \}$.  However, restricting the plan to $\mathcal{G}_v$ may lead to an inefficient path due to the constraint defined in Eq. \ref{eq:setcover_cond2} used for constructing $\mathcal{G}_v$.
For instance, as illustrated in Fig. \ref{fig:tv}, 
replacing  $\tilde{e}_{12}$ with $\mathcal{\tau}_v = \{x_1, x_7, \cdots, x_2\}$ is suboptimal. To search for a more optimal route, we expand $\mathcal{G}_v$ to the roadmap $\mathcal{G}_r$ such that $\mathcal{G}_v \subset \mathcal{G}_r$.
The roadmap is a graph of $\mathcal{G}_r = (\mathcal{V}^*, \mathcal{E}_r)$ where the node set is the same as $\mathcal{V}^*$ and the edge set $\mathcal{E}_r$ satisfies $x_i \mapsto x_j$ for $\forall e_{ij} \in \mathcal{E}_r$ but the sensor-range constant $r$ is a bit relaxed to explore the more free space. $\mathcal{G}_r$ can be efficiently computed using Delaunay triangulation \cite{DT80} with edge cutting.
Fig. \ref{fig:roadmap_graph} shows a roadmap $\mathcal{G}_r$ which has newly added edges ($e_{68}, e_{26}, e_{36}$). The only problem of $\mathcal{G}_r$ is that the additional edges in $\mathcal{G}_r$ may not satisfy the distance constraint $r$, and we need to add a Steiner node in the edges, for instance, the yellow node $x_9$ in Fig. \ref{fig:tr}, between $x_2$ and $x_6$.



Given two route choices $\mathcal{\tau}_r$ and $\mathcal{\tau}_v$, the next challenge is to make a more optimal choice between them. It is important to note that this problem is not simply choosing the one with the shorter distance, as the robot needs to physically visit the node and spend time capturing images at the new place. To compensate for this cost, we have developed the following cost function:
\vspace*{-0.5em}
\begin{equation}\label{eq:replan-weight}
    \psi(\mathcal{\tau})= (1-\eta)\sum_{i}^{m-1} \rho( x_{i+1}, x_i) + \eta \cdot n,
\end{equation}
where $n$ represents the number of newly added viewpoints, while $\eta$ is a penalty factor balancing the total path length and the number of additional viewpoints. We observed that a new viewpoint adds approximately 50 more seconds with our panoramic scanning. 
Lastly, we substitute all $\tilde{e}_{ij} \in \mathcal{T}$ with $min(\psi(\tau_v), \psi(\tau_r))$ to obtain the final scan sequence $\mathcal{T}^*$. Fig. \ref{fig:traj_star} illustrates the final scanning sequence $\mathcal{T}^*=\{e_{17}, e_{76}, e_{69}, \cdots, e_{56}, e_{67}, e_{78} \}$.


After $\mathcal{T}^*$ is obtained, a standard motion planner plans the continuous motion. In our case, we use the A* algorithm for the global planning between successive viewpoints $x_i, x_{i+1}$ and employ the TEB planner~\cite{RosBer17} for the collision-free local planning.

\section{Experiments}\label{section:Experiment}





\subsection{ Performance Analysis in Synthetic Environments }\label{subsection:synthetic_exp}
\begin{figure}[htb!]
\begin{center}
\subfigure[Room]{\label{fig:room-sim}\includegraphics[width=0.25\columnwidth]{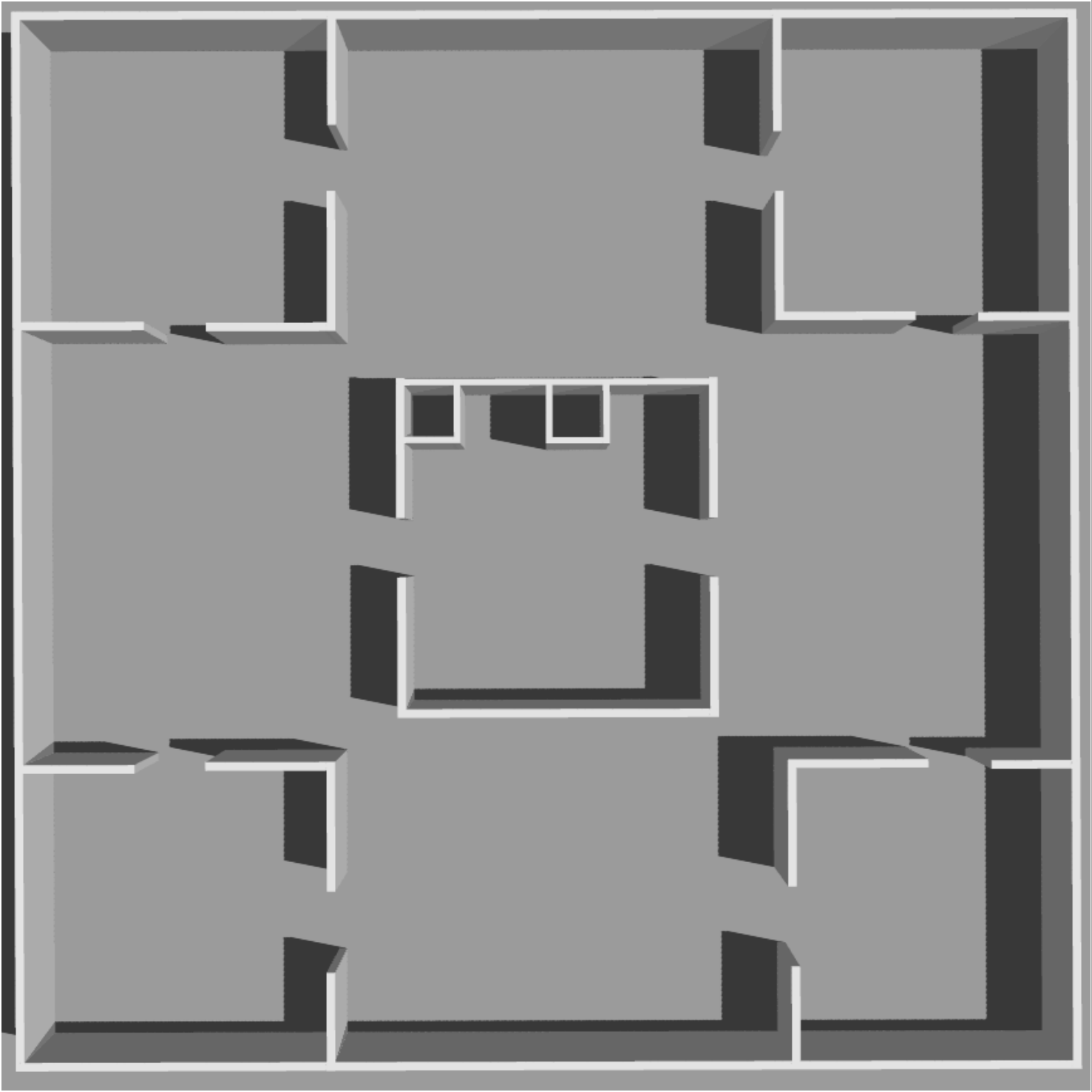}} 
\subfigure[Rooms]{\label{fig:rooms-sim}\includegraphics[width=0.25\columnwidth]{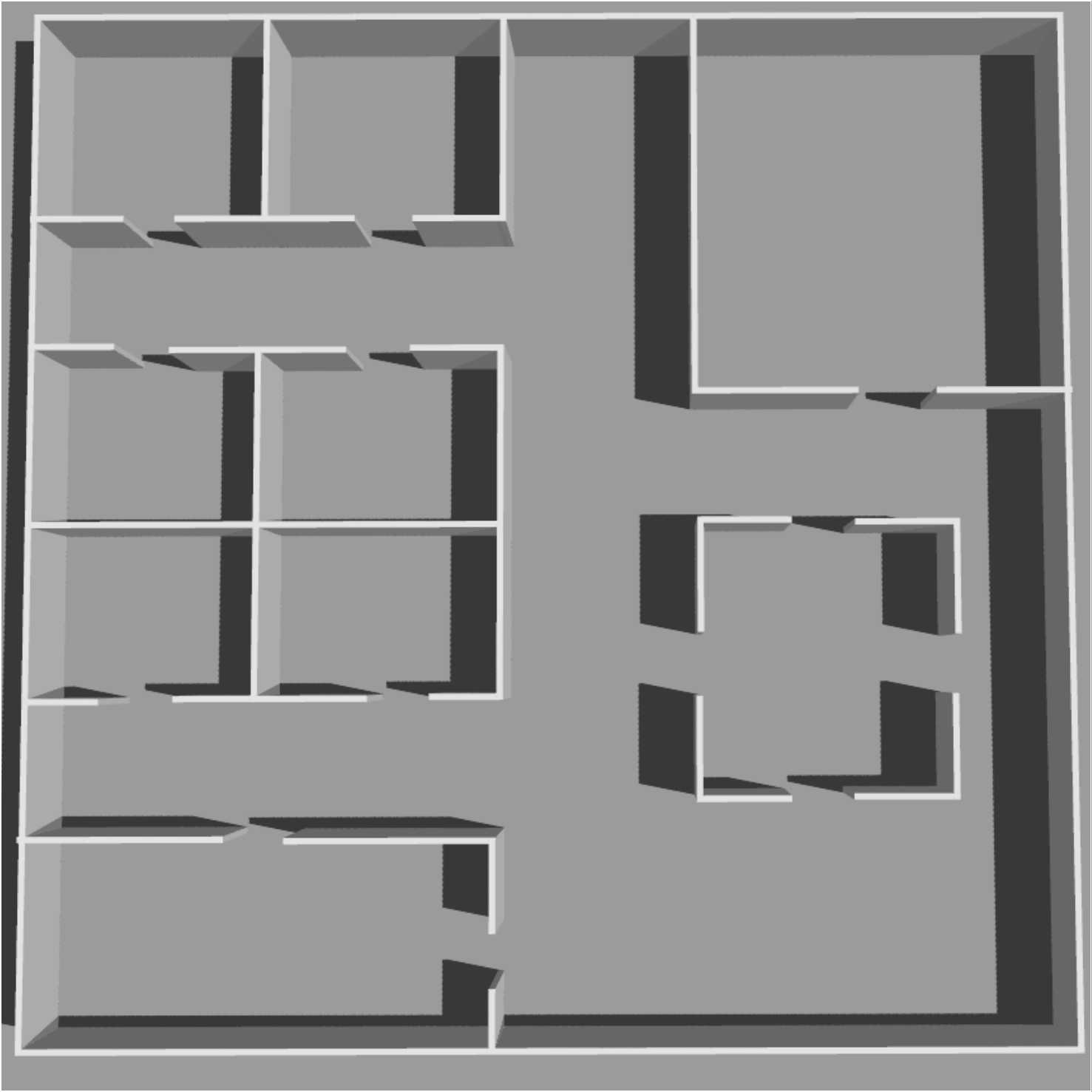}} 
\end{center}
\vspace*{-1.0em}
\caption{\textbf{Examples of synthetic worlds.} The modified "room" and the newly created "rooms" world.}\label{fig:syn_envs}
\vspace*{-1.0em}
\end{figure}

This section compares the performance of our method with other state-of-the-art approaches in synthetic environments. All experiments were conducted on the same machine, equipped with an AMD Ryzen 9 CPU (16 cores), and an NVIDIA RTX 3090 GPU, running Ubuntu 20.04 OS.
For the synthetic world experiments, we utilized Gazebo worlds provided by Explore-Bench \cite{XuYan22}, originally designed for operating the TurtleBot3 robot. However, for this experiment, we made slight modifications to one of the existing worlds, named "room", to accommodate a larger robot, enabling it to pass through the narrow passages of the space. 
We use the testing robot as the Jackal for all experiments. Additionally, we created a custom world, called "rooms", based on the worlds in Explore-Bench, resulting in seven Gazebo worlds including the two new worlds shown in Fig. \ref{fig:syn_envs}.

To ensure a diverse set of baselines, we selected four state-of-the-art methods discussed in Section \ref{sec:Related_work}. Specifically, we included two CPP approaches: Boustrophedon (BCD) \cite{ChoPig98}, contour line-based CPP (CLCPP) \cite{BorHag18}, a global Kriging variance minimization (GKVM) based IPP solver \cite{XiaWac22}, and one active SLAM method: 3DMR \cite{FreRui23}. Notably, \cite{FreRui23} is one of the latest implementations of the next-best-view (NBV) planning approach \cite{BirSie16}.
Since the NBV approach is an online adaptive planner, we first allowed the module to explore the test environment into 2\textit{m} sensor range satisfying the overlapping constraint (Eq. \ref{eq:setcover_cond2}). The viewpoints used for comparison were selected based on those chosen during the online NBV process. 
For the other three baselines, we first utilized Autoexplorer \cite{HanKim22} to construct the base map. Using this map, we applied each method to generate a view plan. The final set of viewpoints was obtained by segmenting these trajectories into 2\textit{m} segments, ensuring compliance with Eq. \ref{eq:setcover_cond2} and \ref{eq:tsp_cond2}.

Table \ref{tab:siml-results-cov-num} summarizes the total coverage achieved and the number of viewpoints generated by each approach. Table \ref{tab:siml-results-time} reports both exploration and planning times. 
Fig. \ref{fig:sim-results} shows an example of the "rooms" environment where we qualitatively compare our method against baseline approaches. GKVM achieved the shortest path length and the lowest number of viewpoints; however, it only covered approximately 74\% of the space. The two CPP methods attained full 100\% coverage but required a significantly higher number of viewpoints, leading to inefficient path lengths. In contrast, our method achieved over 99\% coverage while requiring only 133 viewpoints, demonstrating a balance between efficiency and completeness.

According to our study, CPP approaches, such as the BCD method, exhibit the fastest planning time and the highest coverage among the five methods. However, BCD requires nearly five times as many viewpoints as our method. 
Similarly, CLCPP demonstrates relatively short planning times and high coverage. However, like BCD, it also requires significantly more viewpoints than our method.
On the other hand, GKVM achieves the lowest number of viewpoints among all approaches. However, its coverage is significantly lower than other methods, and in some cases, it fails to complete the input map reliably.
Lastly, 3DMR finds a balance between coverage and the number of viewpoints. However, it is about five times slower than other methods regarding total planning time, including view planning time and space exploration time. This is primarily because the next-best-view type planners are better suited for unknown space exploration rather than for selecting viewpoints in a post-processing scenario.

Overall, our method generates the least number of viewpoints while achieving over 99\% 
scanning coverage. This is because only our method incorporates the visibility constraint during the optimization of the view plan, whereas other approaches do not consider such a constraint. 
The two CPP-based methods achieved faster planning times than our approach. However, the planning time constitutes a relatively small portion of the total scanning time, which includes planning, navigation, and image capturing. This is because the number of viewpoints primarily influences the total scanning time. In Sec.~\ref{subsection:realworld_exp}, we will further discuss the comparison of total scanning time between our approach and other baseline methods.

\subsection{ Real World Experiments }\label{subsection:realworld_exp}


In this section, we conducted real-robot experiments to assess the total scanning time, encompassing the planning, the robot's motion, and the image capture time. Since the total scanning time is closely associated with planning efficiency, the real-world experiment provides a comprehensive analysis of the proposed approach compared to baseline methods. For this experiment, we excluded the 3DMR approach from the four methods discussed in \ref{subsection:synthetic_exp} due to its requirement for a 3D LiDAR sensor.

\begin{figure}[H]
\begin{centering}
{\includegraphics[width=0.9\columnwidth]{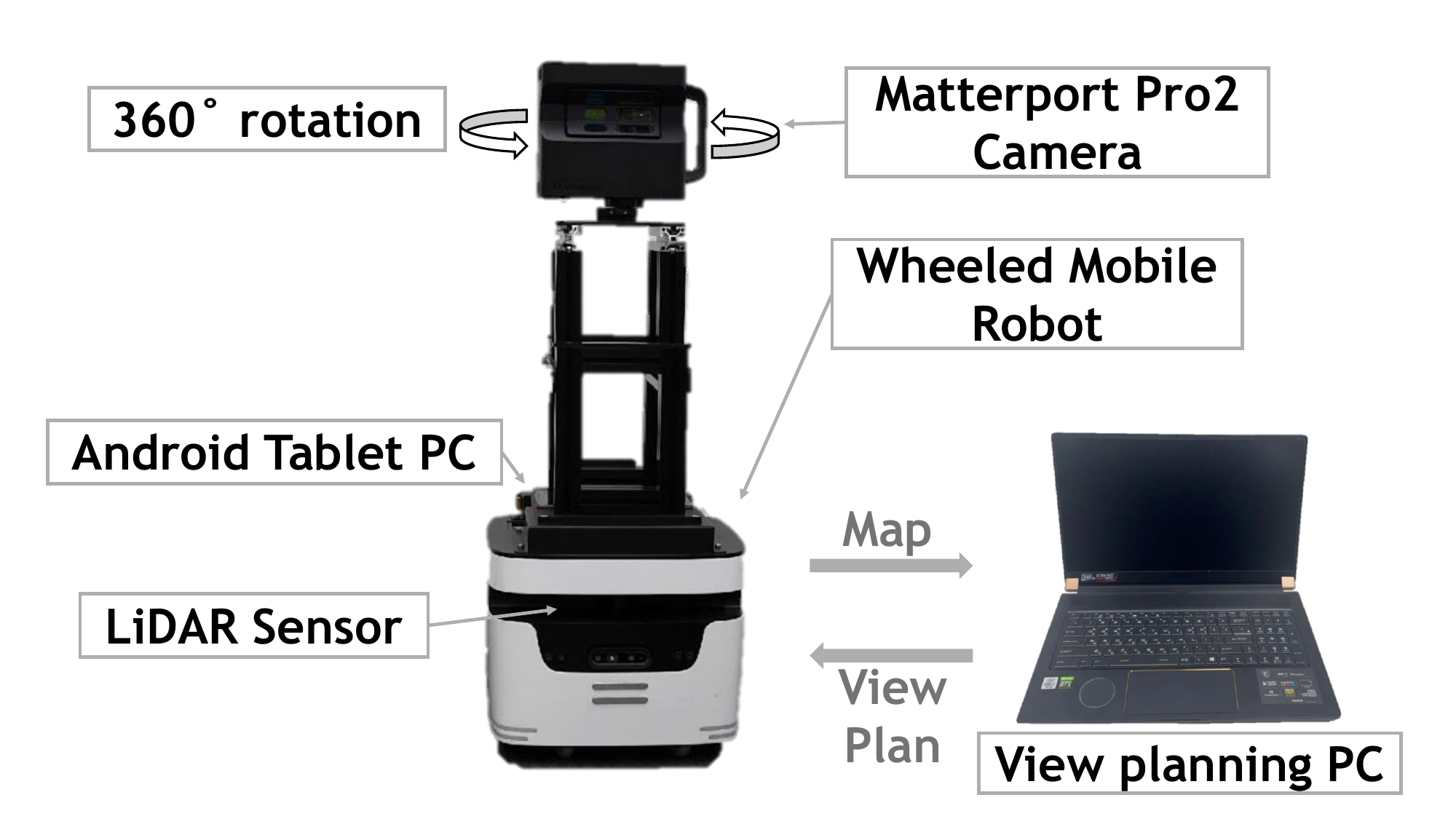}}
\captionof{figure}{ \textbf{Robot platform for real-world experiments.} } \label{fig:robot}
\end{centering}
\vspace*{-0.5em}
\end{figure}

Fig. \ref{fig:robot} introduces our scanning-bot system utilized for real-world experiments. The system consists of a view planning PC and a Matterport Pro2 panoramic camera mounted on top of a differential-wheeled mobile robot. This robot is equipped with basic sensors for 2D LiDAR SLAM. It is powered by an Intel i5 processor with 8GB of RAM, enabling on-board navigation processes such as SLAM and motion planning. The view planning PC has an Intel Core i9 CPU and an NVIDIA RTX 3080 GPU, running Ubuntu 18.04. 
The captured images are transmitted to Matterport's Cortex AI platform to reconstruct a textured 3D mesh by \cite{ChaZha17}. A crucial constraint for Matterport reconstruction is that each scanning point should be within the 2$m$ range of neighboring scanning positions.
\vspace*{-0.5em}

{\begin{table*}[htb]
\setlength{\tabcolsep}{10pt}
\renewcommand{\arraystretch}{1.1}
\begin{centering}
{$
\begin{array}{@{}l*{7}{c}@{}}

\toprule[1.5pt]
 \multirow{3}{*}{Methods} & \multicolumn{7}{c@{}}{\text{ Coverage (\%) / Number of viewpoints }} \\
    \cmidrule(l){2-8}
    & \text{corner} & \text{corridor} & \text{loop} & \text{loop\_ with\_ corridor} & \text{room} & \text{rooms} & \text{room\_ with\_ corner} \\ 
\midrule
\text{BCD} & {99.8 / 259 } & {\textbf{100} / 304 } & {\textbf{100} / 219 } & {\textbf{100} / 404}  & {\textbf{100} / 594} & {\textbf{100} / 525 } & {\textbf{100} / 342 }\\ 
\text{CLCPP} & {\textbf{100} / 296 } &  {\textbf{100}  / 256 } &  {\textbf{100}  / 191 } &  {\textbf{100}  / 277 }  &  {\textbf{100}  / 480 } & {\textbf{100}  / 516 } &  {99.9  / 386} \\
\text{GKVM} &  {\;\;-\;\;  / \;\;-\;\; } &  {85.7  / \textbf{48} } &  {\;\;-\;\;  / \;\;-\;\; } &  {87.4  / \textbf{60} }  &  {72.3  / \textbf{72} } & {74.2  / \textbf{94} } &  {72.2  / \textbf{52} }\\ 
\text{3DMR} &  {99.3  / 86 } &  {99.3  / 75 } &  {99.9  / 61 } &  {99.3  / 93 }  &  {99.4  / 182 } & {99.1  / 170 } &  {99.4  / 117 }\\
\text{Our method} &  {99.0 / \textbf{58} } &  {99.4  / 70 } &  {99.6  / \textbf{36} } &  {99.9  / 70 }  &  {99.6  / 127 } & {99.5  / 133 } &  {98.9  / 89 }\\ 
\bottomrule[1.5pt]
\end{array} 
$}
\caption{ \textbf{The quantitative comparisons of coverage and number of viewpoints for different methods in various simulation worlds.} "$-$" indicates that no feasible plan was generated within a finite time.}\label{tab:siml-results-cov-num}
\end{centering} 
\end{table*}}

\begin{table*}[htb]
\renewcommand{\arraystretch}{1.1}
\begin{centering}
{$
\begin{array}{@{}l*{7}{c}@{}}

\toprule[1.5pt]
 \multirow{3}{*}{Methods} & \multicolumn{7}{c@{}}{\text{ Exploration time (s) / Planning time (s) }} \\
    \cmidrule(l){2-8}
    & \text{corner} & \text{corridor} & \text{loop} & \text{loop\_ with\_ corridor} & \text{room} & \text{rooms} & \text{room\_ with\_ corner}\\ 
\midrule
\text{BCD} &  356.12  / \textbf{0.01} &  207.44 /\textbf{0.01} &  258.98 /\textbf{0.01} & 661.62 /\textbf{0.01}  & 360.58 /\textbf{0.02} & 314.39 /\textbf{0.01} &  347.47 /\textbf{0.01}\\ 
\text{CLCPP} & 356.12  / \textbf{0.01} & 207.44 /\textbf{0.01} &  258.98 /\textbf{0.01} &  661.62 /\textbf{0.01}  &  360.58 /0.11 & 314.39 /0.04 &  347.47 /0.07\\ 
\text{GKVM} &  \;\;-\;\; &  207.44 /17.56 &  \;\;-\;\; &  661.62 /15.27  & 360.58 /12.50 & 314.39 /12.23 &  347.47 /10.39\\ 
\text{3DMR} &  2843.07 &  2246.83 &  1366.92 &  3572.30  &  9180.33 & 6939.96 &  5256.27\\
\text{Our method} & 356.12  / 7.82 & 207.44 /7.57 & 258.98 /5.62 &  661.62 /8.60  &  360.58 /65.68 & 314.39 /63.02 &  347.47 /10.37\\ 
\bottomrule[1.5pt]
\end{array} 
$}
\caption{ \textbf{Comparison of exploration and planning times with baseline methods in sim environments.} The exploration time is identical for BCD, CLCPP, GKVM, and our method, as they all use the map generated by AutoExplorer \cite{HanKim22}.}\label{tab:siml-results-time}
\end{centering} 
\end{table*}

\begin{table*}
\begin{tabular}{c | c | c | c | c }

BCD & CLCPP & GKVM & 3DMR & Our method \\
\makecell{\includegraphics[width=0.37\columnwidth]{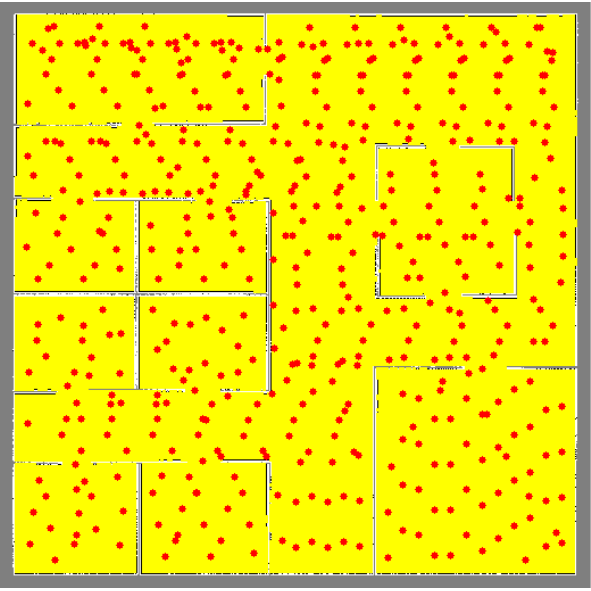} \\ \# VPs: 525, Cov: 100\% } & \makecell{\includegraphics[width=0.37\columnwidth]{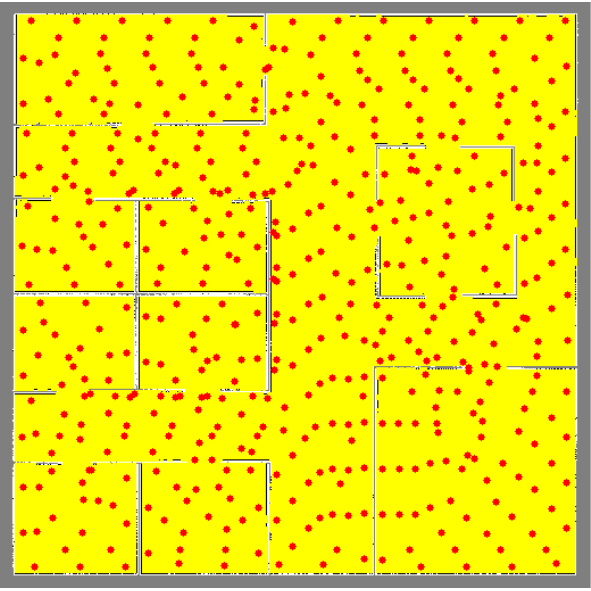} \\ \# VPs: 516, Cov: 100\% } & 
\makecell{\includegraphics[width=0.37\columnwidth]{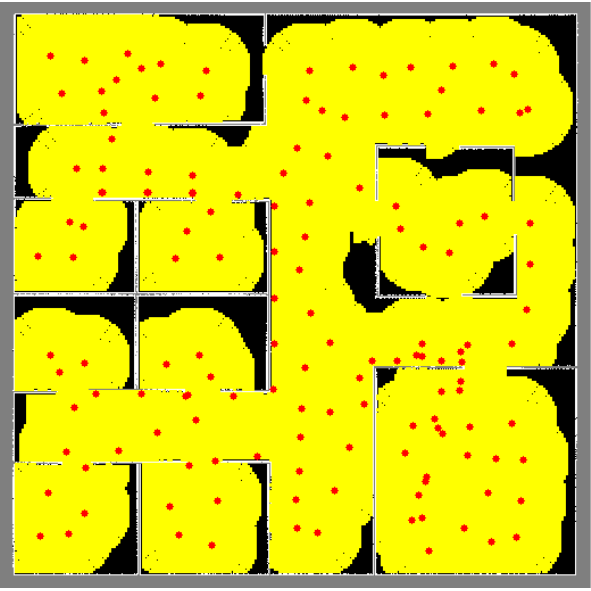}  \\ \# VPs: 94, Cov: 72.3\% } &
\makecell{\includegraphics[width=0.37\columnwidth]{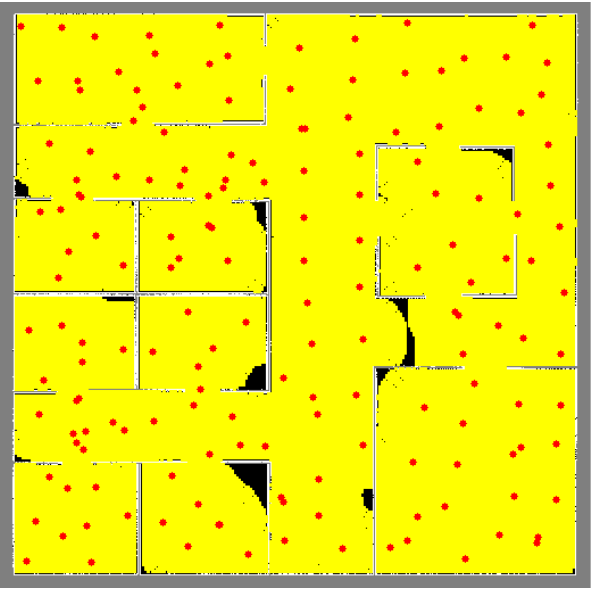} \\ \# VPs: 170, Cov: 99.4\% } &
\makecell{\includegraphics[width=0.37\columnwidth]{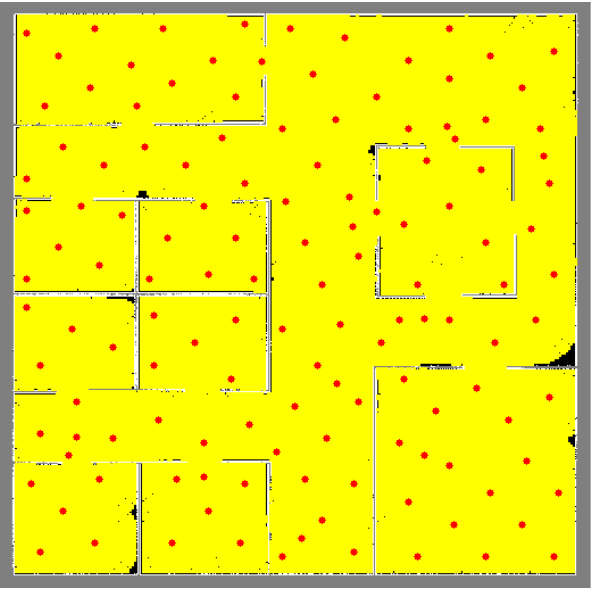} \\ \# VPs: 133, Cov: 99.6\% } \\

\makecell{\includegraphics[width=0.37\columnwidth]{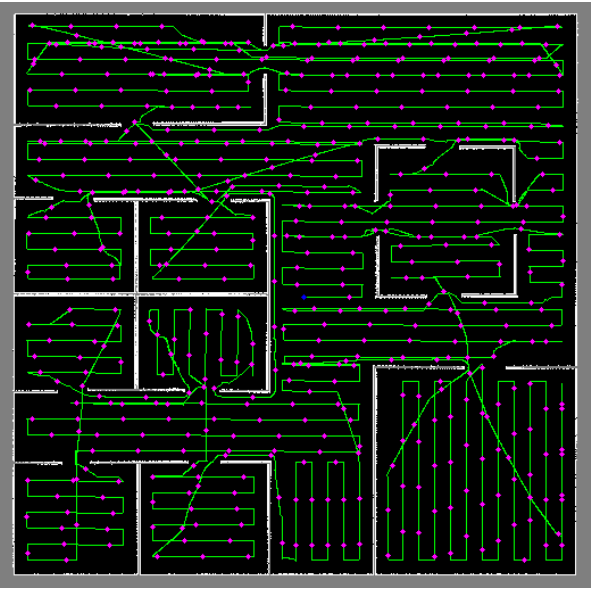} \\ Path length: $1054.74m$ } &
\makecell{\label{fig:sim-icra7-path}\includegraphics[width=0.37\columnwidth]{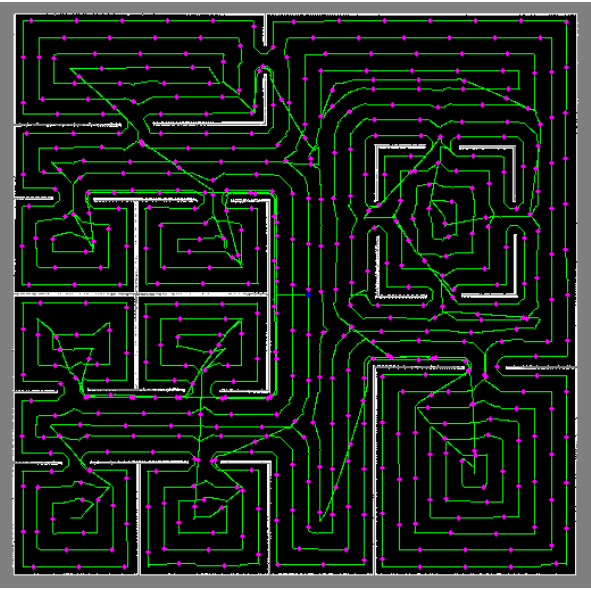} \\ Path length: $1038.24m$ } &
\makecell{\label{fig:sim-ral-path}\includegraphics[width=0.37\columnwidth]{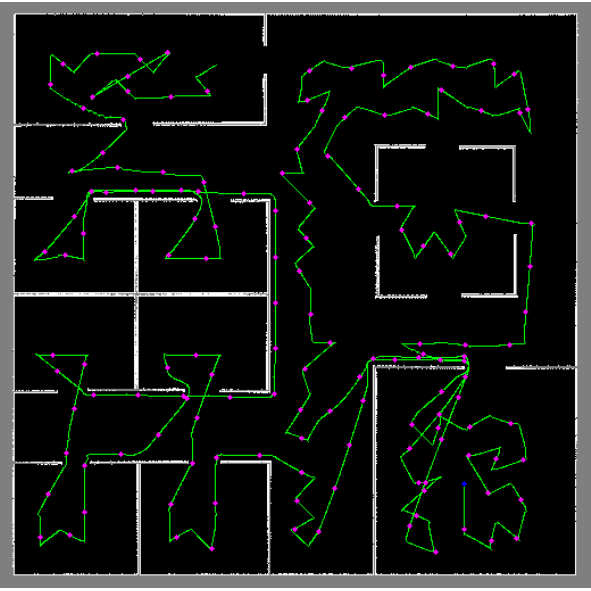} \\ Path length: $191.00m$ }  &
\makecell{\label{fig:sim-corr-path}\includegraphics[width=0.37\columnwidth]{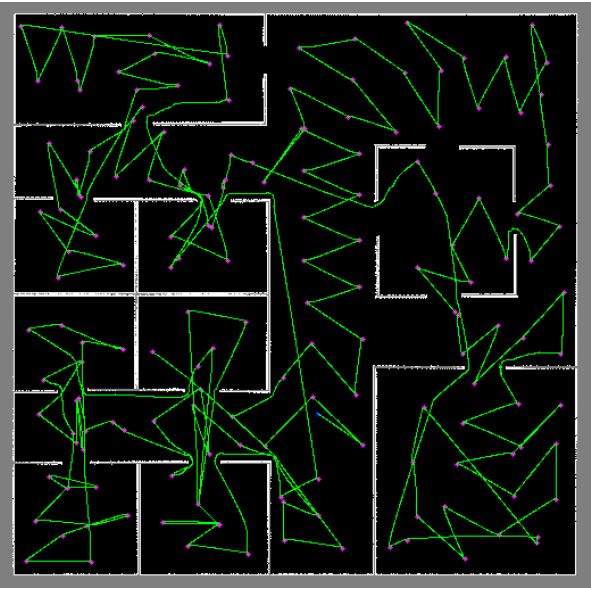} \\ Path length: $508.14m$ } &
\makecell{\label{fig:sim-our-path}\includegraphics[width=0.37\columnwidth]{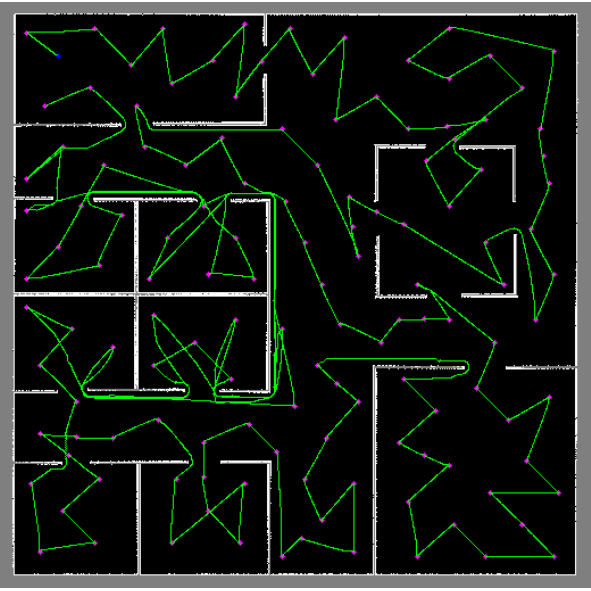} \\ Path length: $428.25m$ }
\end{tabular}

\captionof{figure}{ \textbf{Qualitative comparisons of the sim results}  Top row shows the selected viewpoints (red dots) and the covered area (yellow pixels). The bottom row shows the planned path (green lines). }\label{fig:sim-results}
\end{table*}

\begin{table*}
\begin{tabular}{c | c | c | c }
BCD & CLCPP & GKVM & Our method \\
\makecell{\includegraphics[width=0.48\columnwidth]{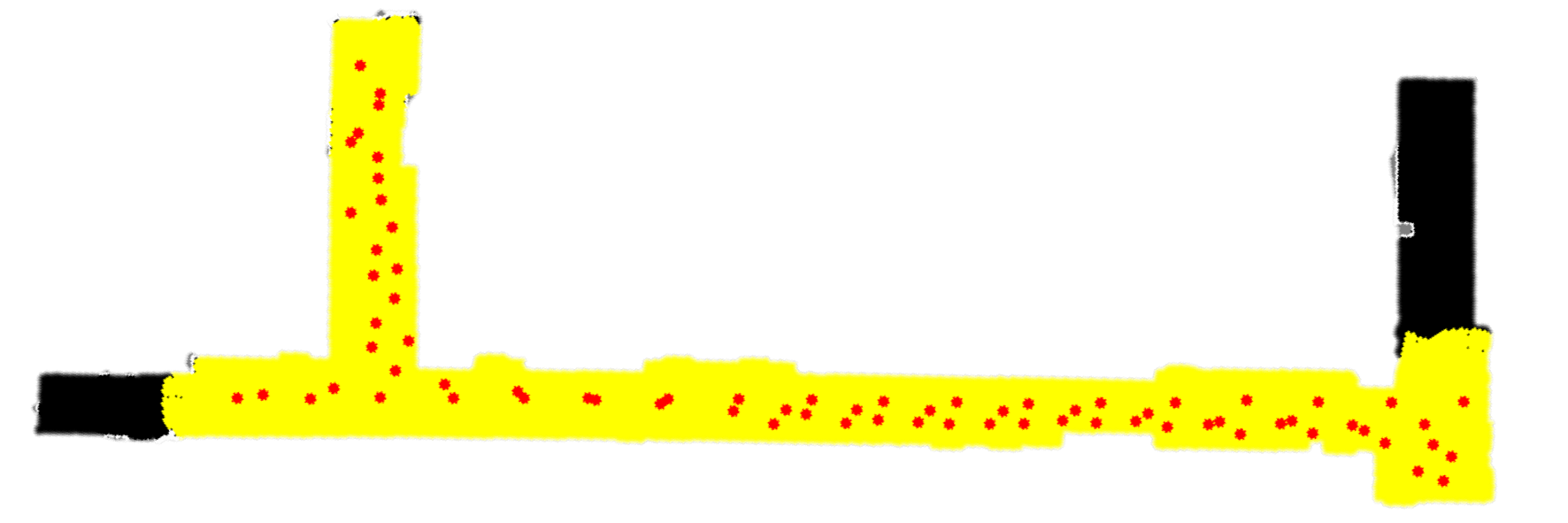} \\ \# VPs: 75, Cov: 83.0\%} & \makecell{\includegraphics[width=0.48\columnwidth]{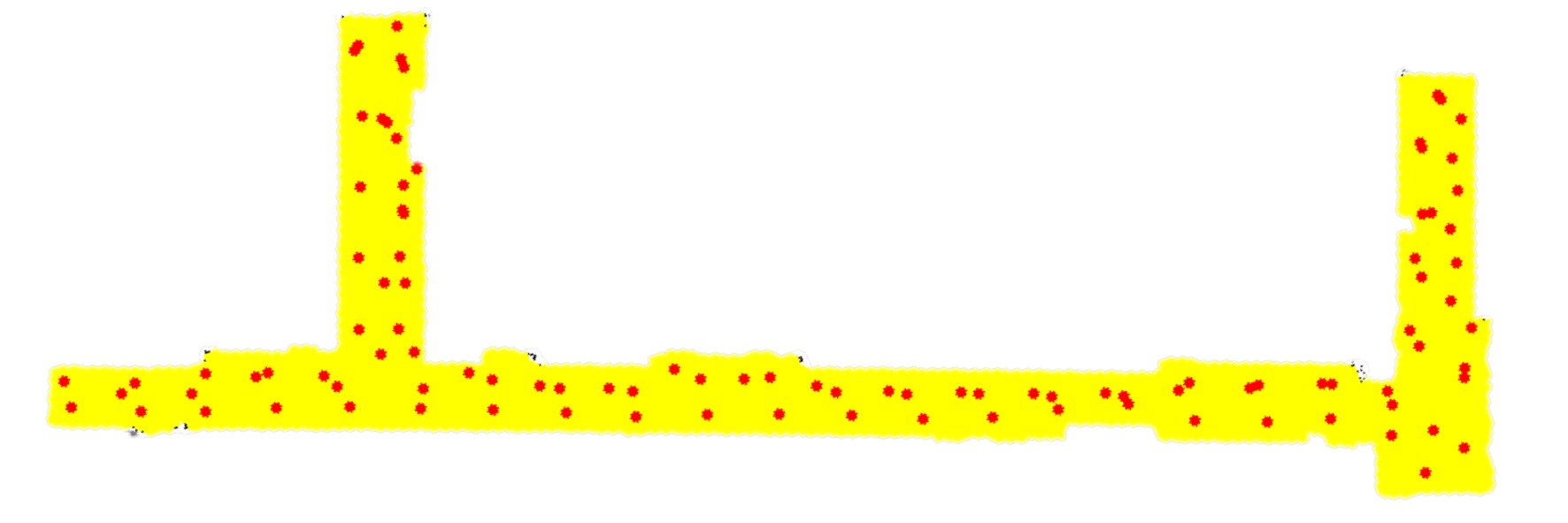} \\ \# VPs: 102, Cov: 100\%} & 
\makecell{\includegraphics[width=0.48\columnwidth]{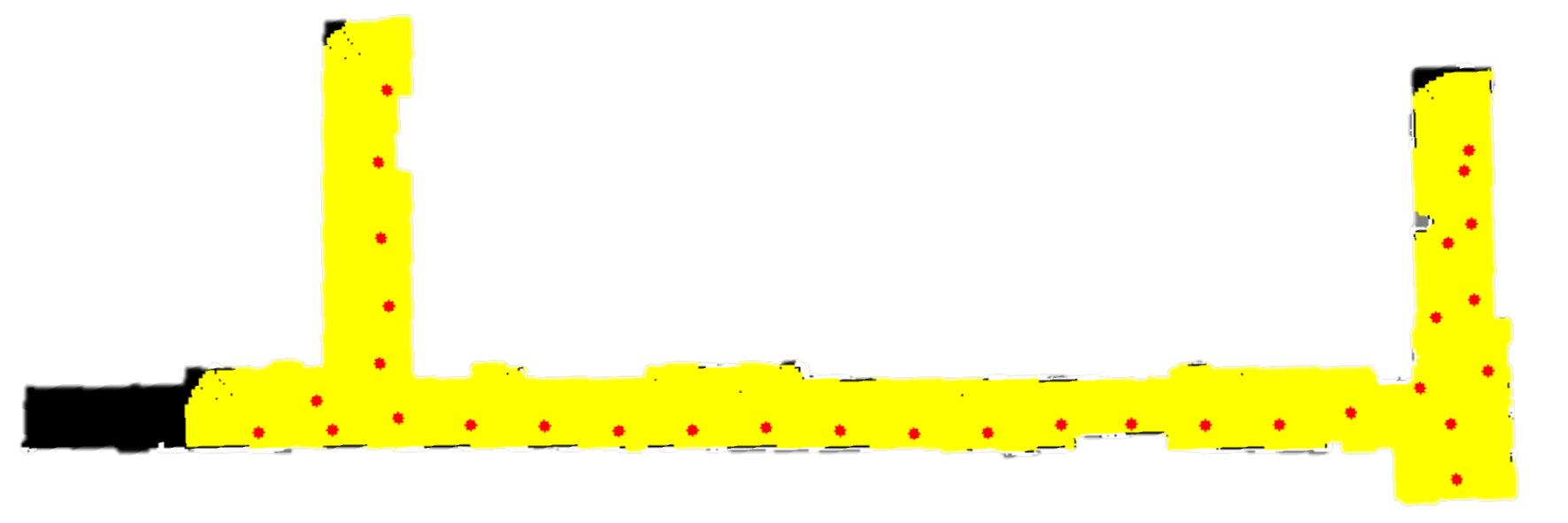} \\ \# VPs: 32, Cov: 93.2\%} &
\makecell{\includegraphics[width=0.48\columnwidth]{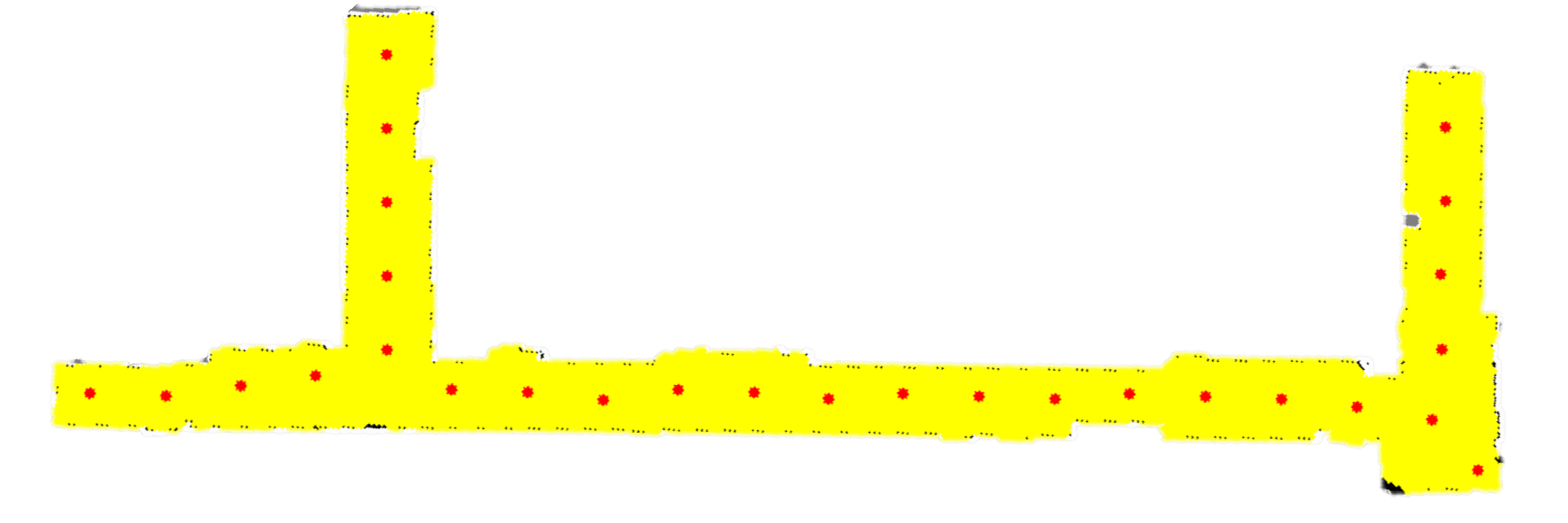} \\ \# VPs: 28, Cov: 99.8\%} \\

\makecell{\includegraphics[width=0.48\columnwidth]{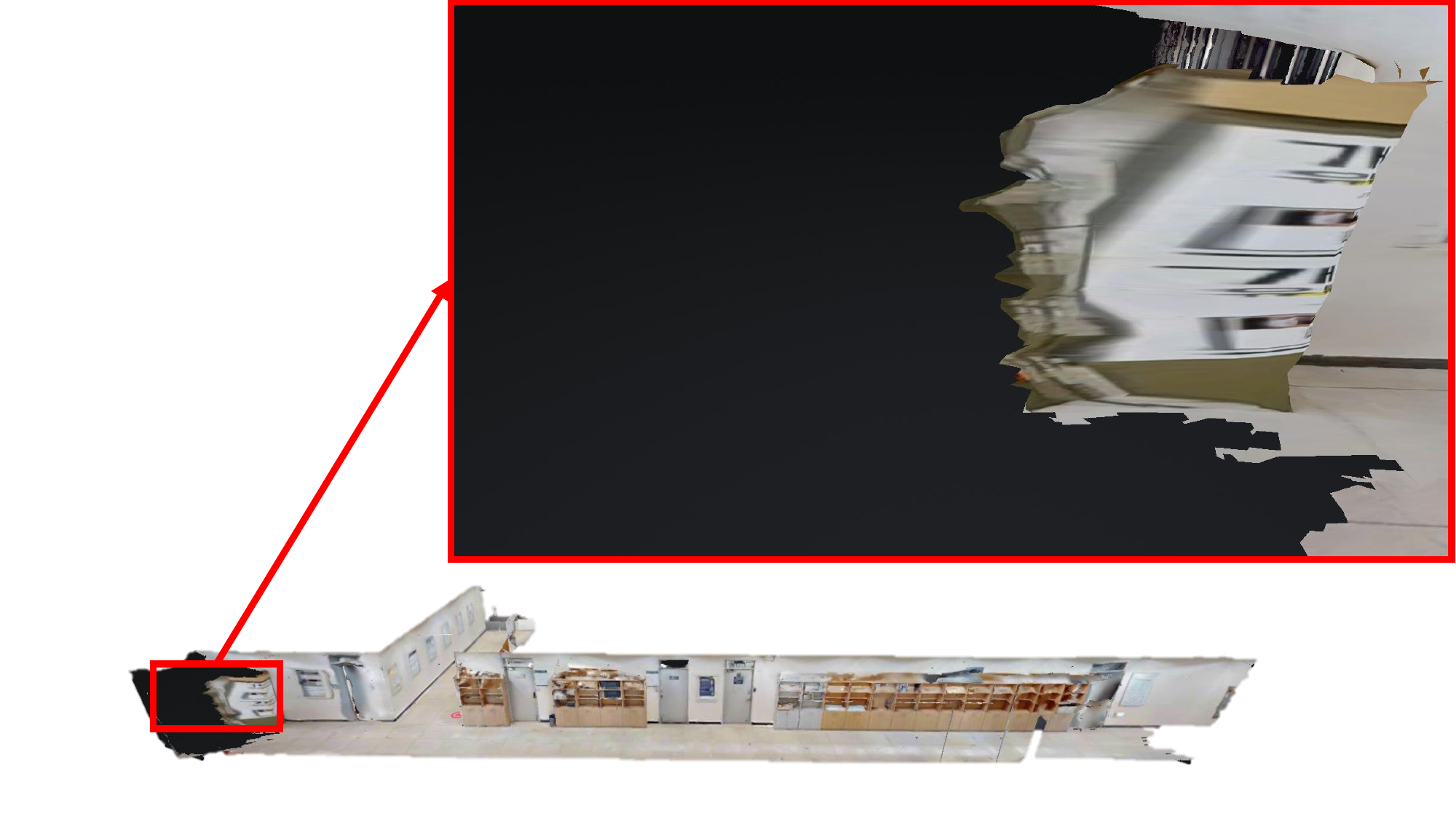} } &
\makecell{\includegraphics[width=0.48\columnwidth]{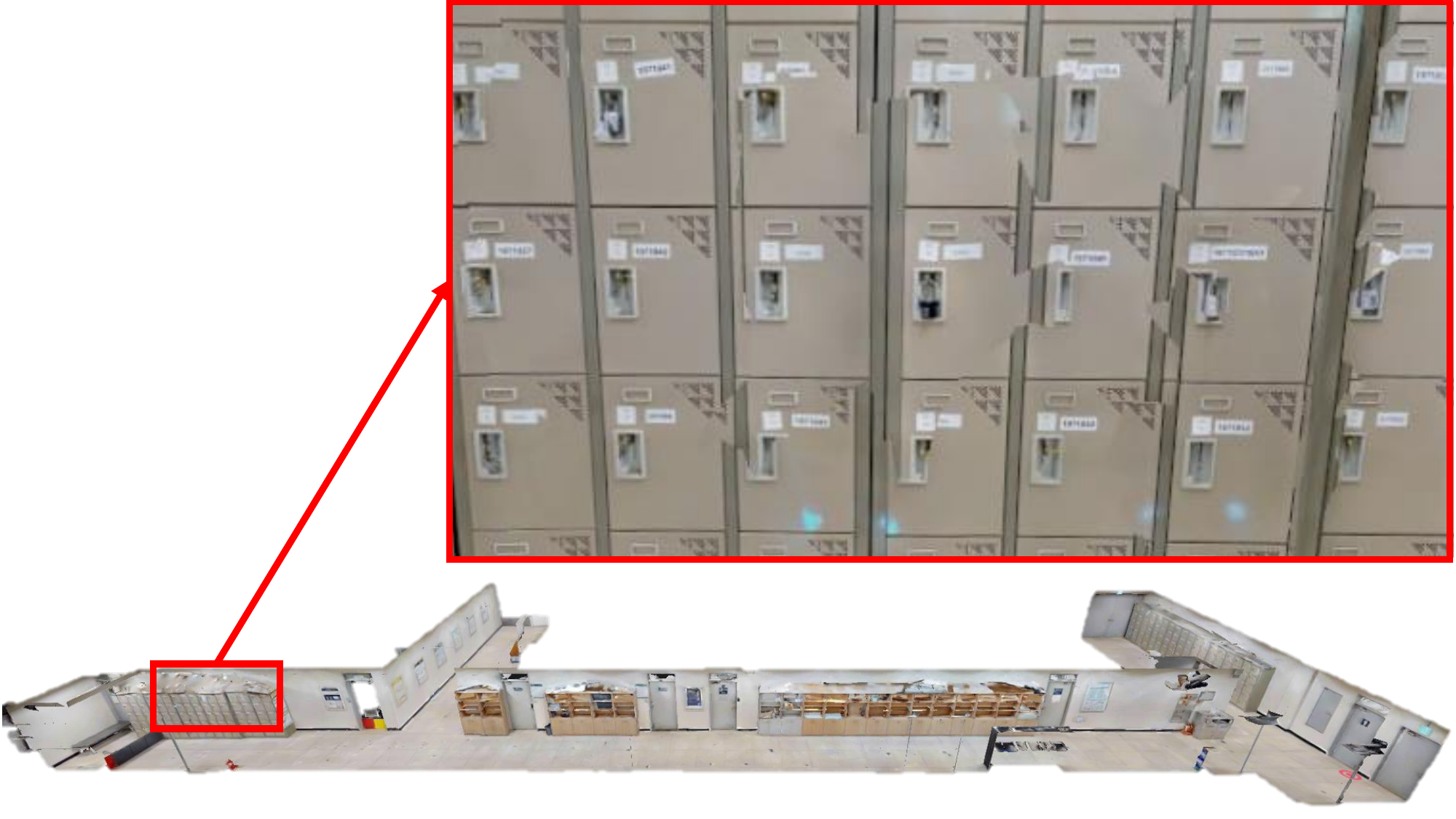} } &
\makecell{\includegraphics[width=0.48\columnwidth]{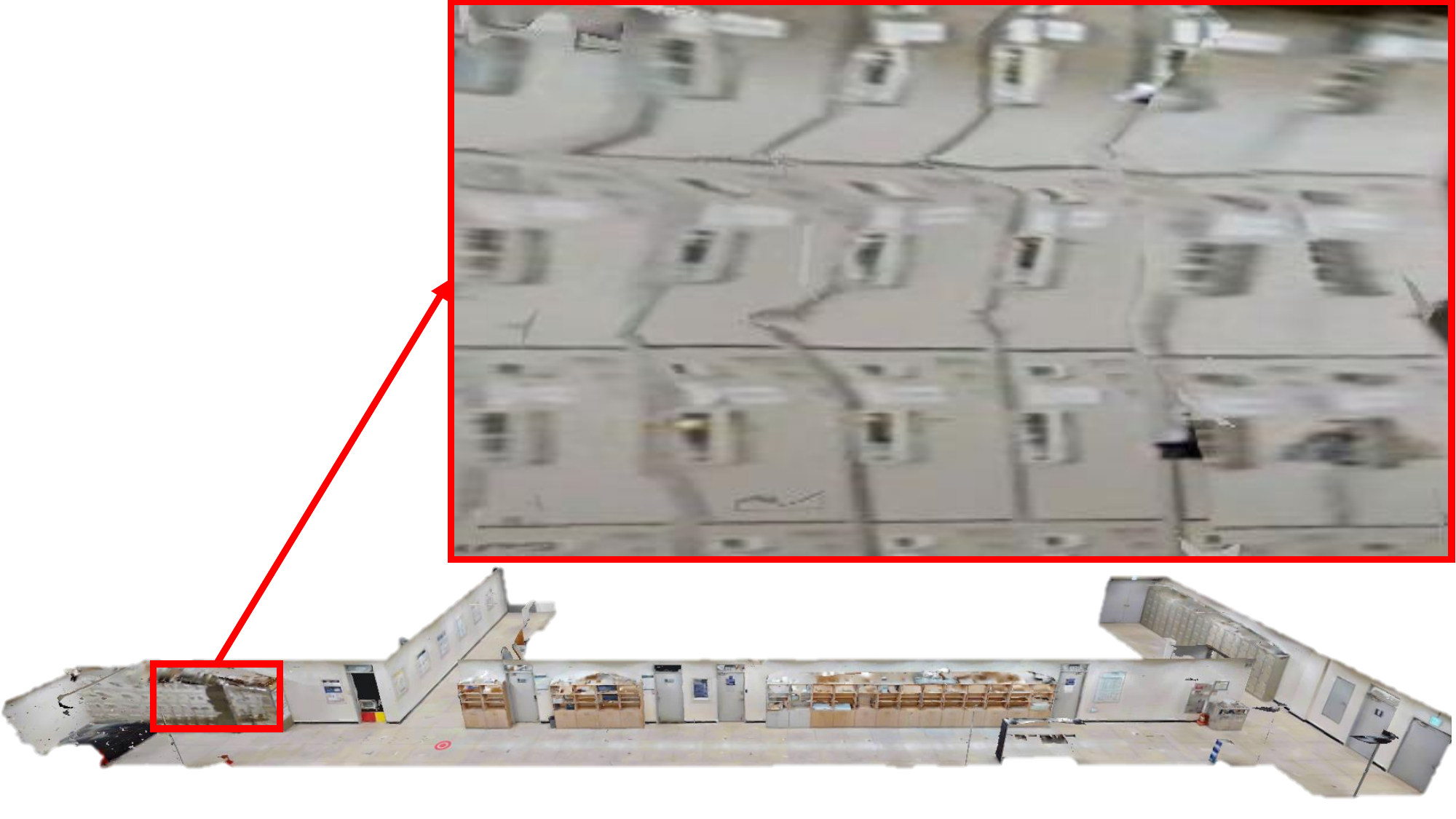} }  &
\makecell{\includegraphics[width=0.48\columnwidth]{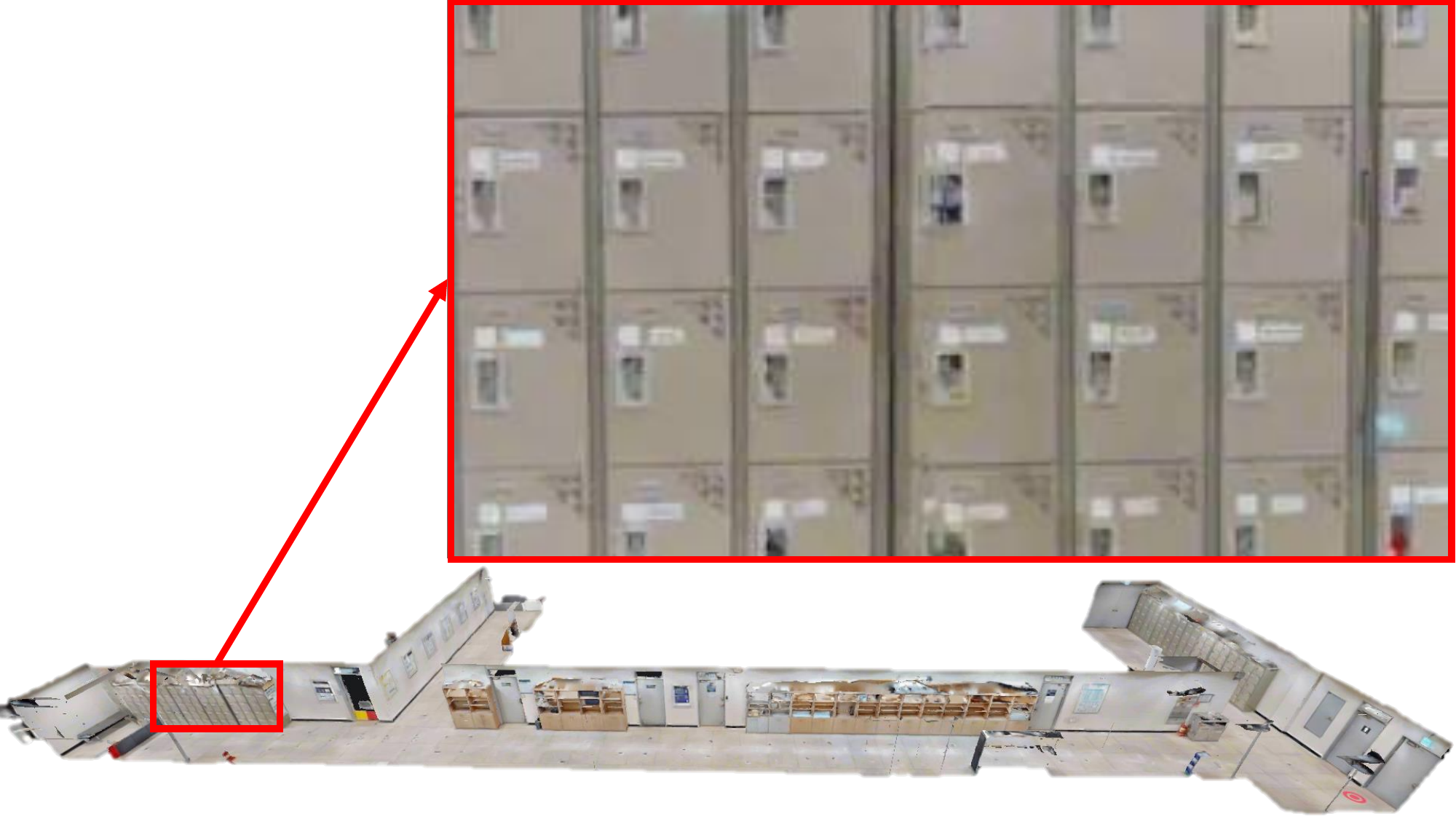} } 
\end{tabular}

\captionof{figure}{\textbf{Qualitative comparisons of real-world experiments.} The top row shows viewpoints (red) and covered areas (yellow). The bottom row shows 3D models with detailed textures, highlighting our method's better-textured mesh quality.}\label{fig:prop-l}
\end{table*}

\begin{table}[htb!]
\setlength{\tabcolsep}{7pt}
\renewcommand{\arraystretch}{1.1}
\begin{centering}
{$
\begin{tabular}{c c c c c}
\toprule[1.5pt]
\text{Methods} & \text{BCD} & \text{CLCPP} & \text{GKVM} & \text{Our method} \\ 
\midrule
\text{Path length ($m$)} & {151.56} & {205.73} & \textbf{65.20} & {112.08} \\ 
\text{Coverage (\%)} & {82.98} & \textbf{100} & {93.21} & {99.81} \\
\text{Number of Viewpoints} & {75} & {102} & {32} &  \textbf{28} \\
\text{Planning time (s)} &  \textbf{0.02} & \textbf{0.02} & {65.24} & {12.85}  \\ 
\text{Scanning time (m)} &  {41} & {95} & {40} & \textbf{33}  \\
\bottomrule[1.5pt]
\end{tabular} 
$}
\caption{ \textbf{Comparisons of real-world results.} Our method achieved over 99\% coverage with the fewest viewpoints and shortest scanning time. }\label{tab:real-results}
\vspace*{-1.0em}
\end{centering}
\end{table}


Similar to Section \ref{subsection:synthetic_exp}, we conducted several experiments to assess the baseline approaches except 3DMR owing to its strict dependency on a 3D LiDAR sensor. Consequently, we report the BCD, CLCPP, and GKVM results as the baselines against our method for these experiments. The experiments were carried out using the same scanning-bot system presented in Fig. \ref{fig:robot}, including the same ROS navigation SW packages \cite{EppKon10}\cite{RosBer17}\cite{Mac21}. 

The experiments were conducted in the ASAN Engineering Building at Ewha Womans University, which consists of a long corridor; see Fig. \ref{fig:prop-l}. Table \ref{tab:real-results} presents the quantitative results of these experiments. We observed that our method and CLCPP achieved over 99\% coverage of the scanning space. However, our method was approximately three times faster than CLCPP in terms of total scanning time. The total scanning time encompasses visiting all planned viewpoints and capturing 3D images at each point.

Fig. \ref{fig:prop-l} highlights the superiority of our method in terms of coverage rate and the number of viewpoints generated. Our method produces fewer viewpoints while ensuring the selected viewpoints thoroughly cover the entire space. 

\begin{figure}[htb!]
\begin{centering}

\subfigure[New Engineering Building]
{\label{fig:eng1}\includegraphics[width=0.95\columnwidth]{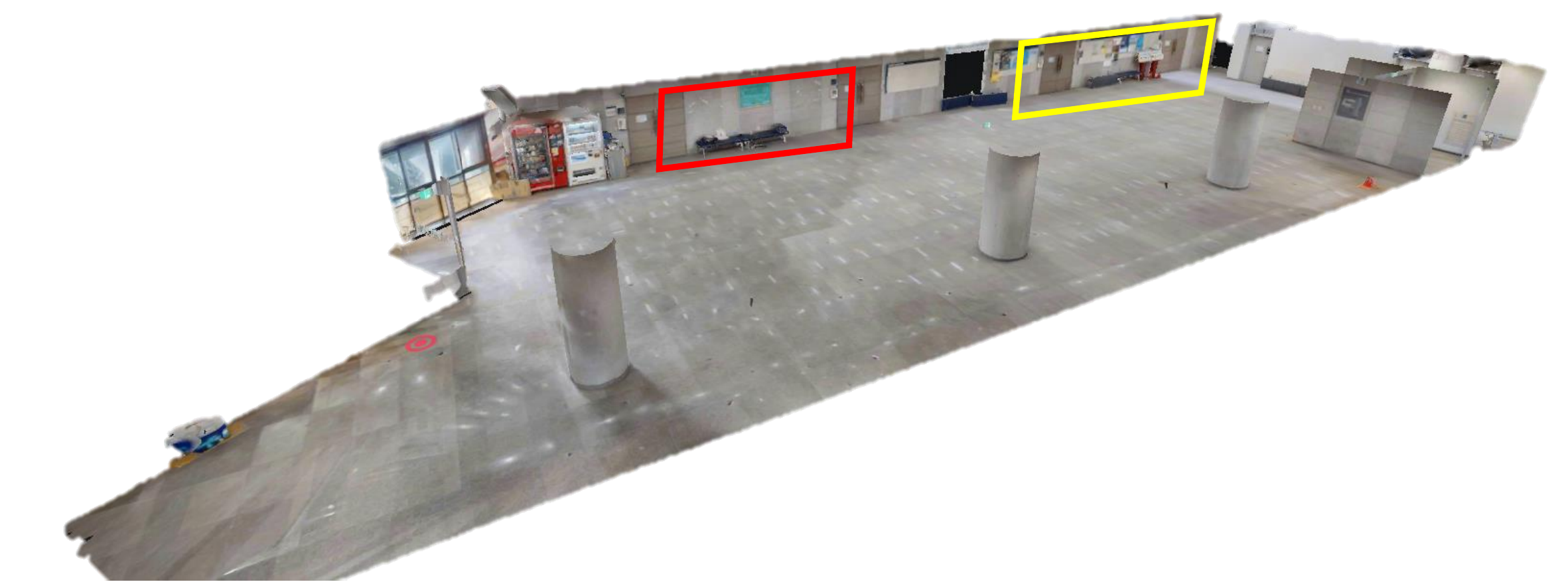}}


\subfigure[Zoomed-in view 1]
{\label{fig:eng1-1}\includegraphics[width=0.44\columnwidth]{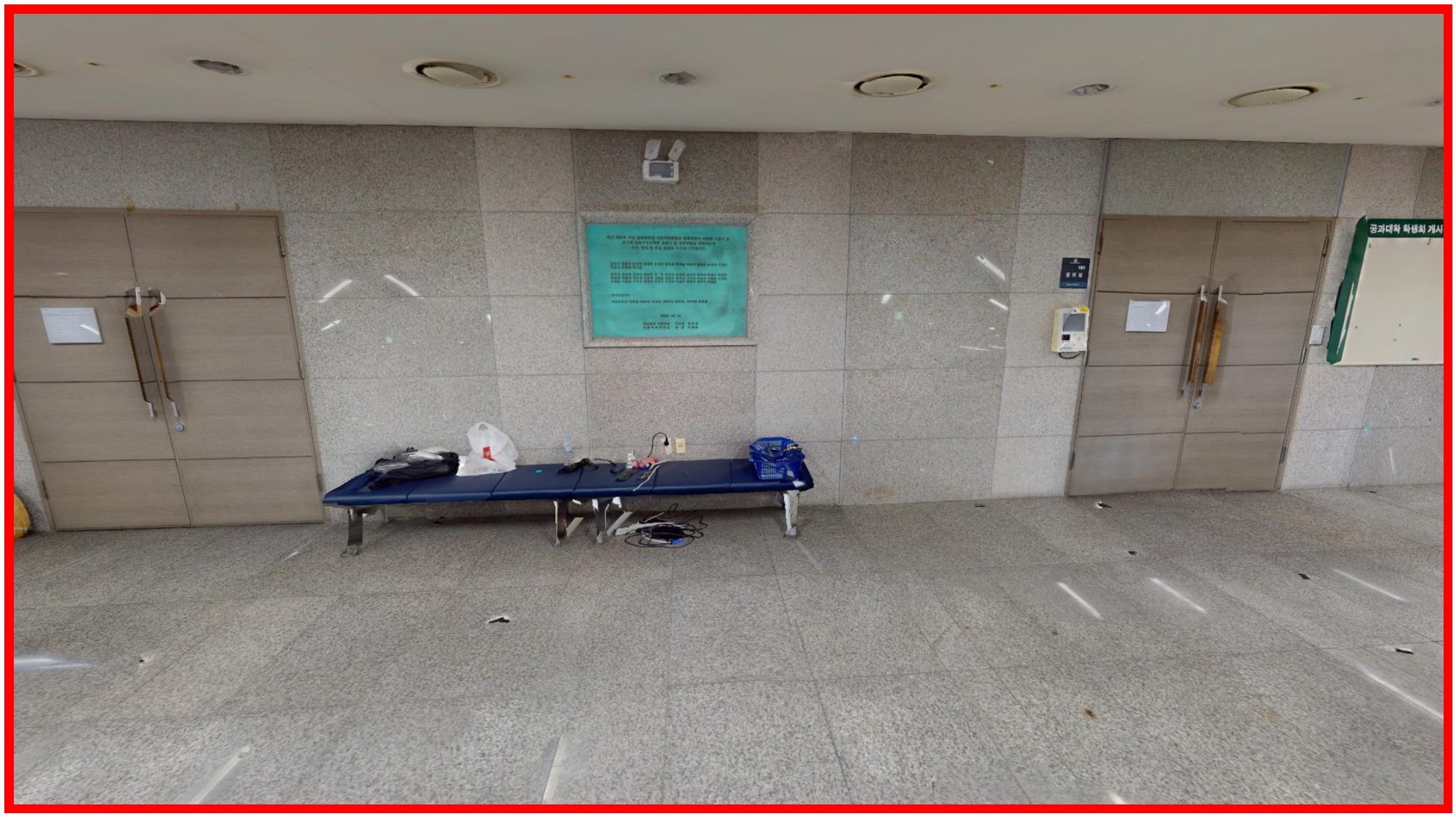}}
\subfigure[Zoomed-in view 2]
{\label{fig:eng1-2}\includegraphics[width=0.44\columnwidth]{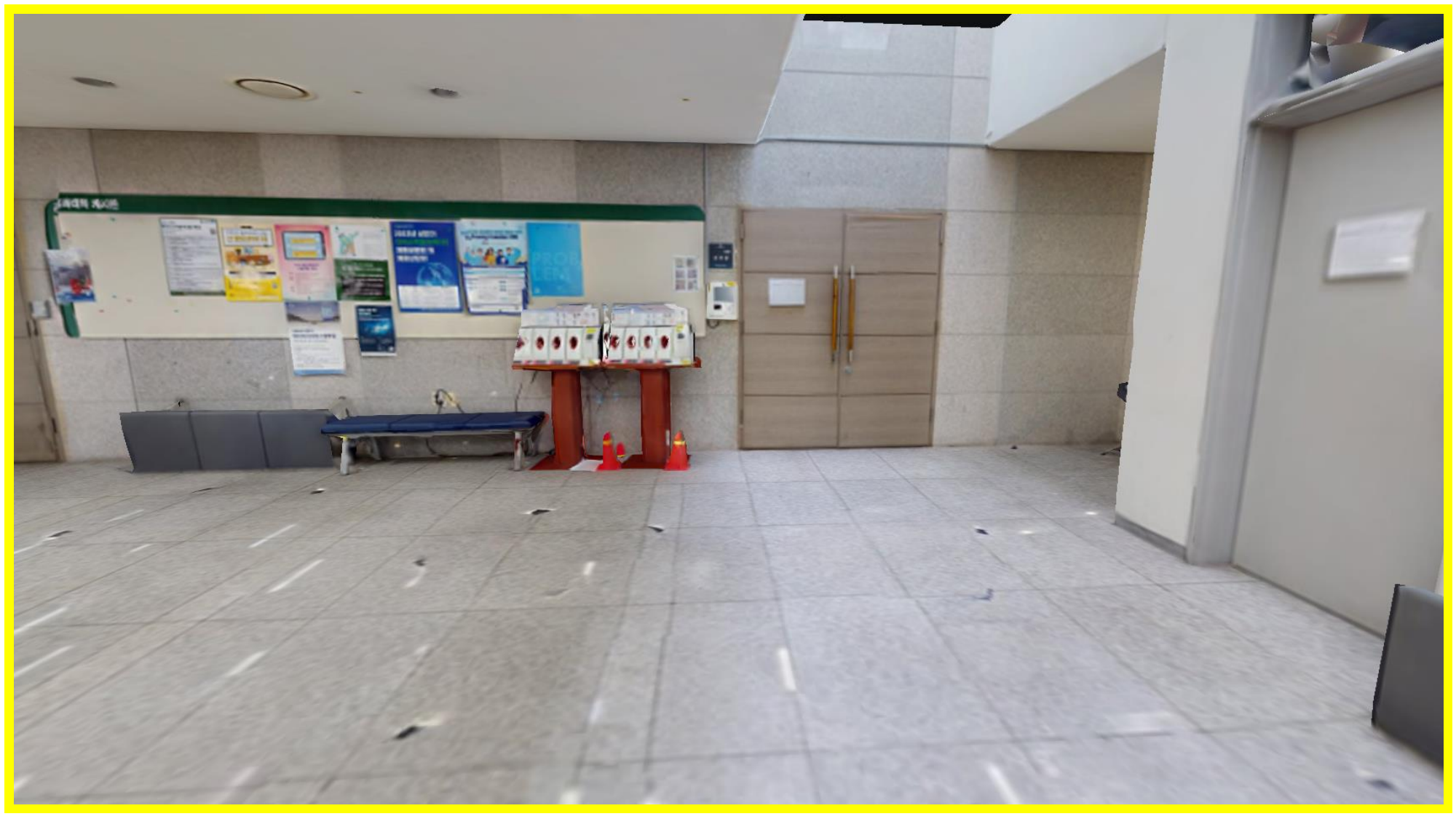}}

\subfigure[SK Telecoms Building]
{\label{fig:enga}\includegraphics[width=0.95\columnwidth]{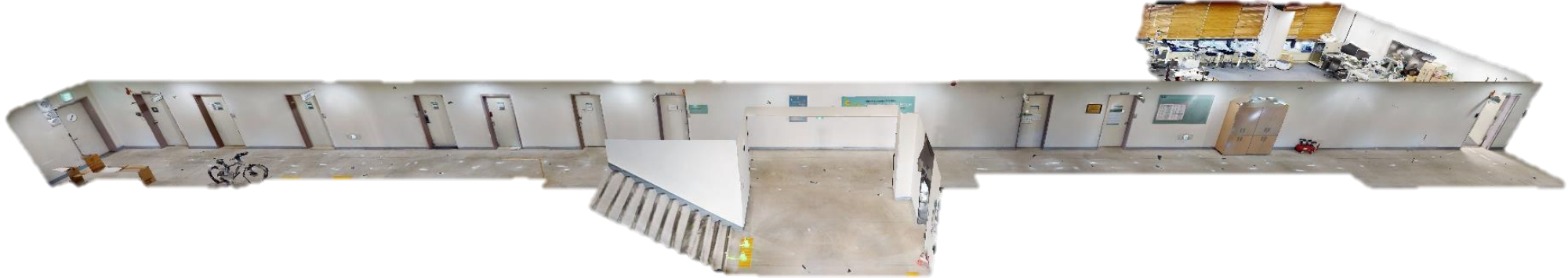}}

\subfigure[Arts \& Design Building]
{\label{fig:art}\includegraphics[width=0.45\columnwidth]{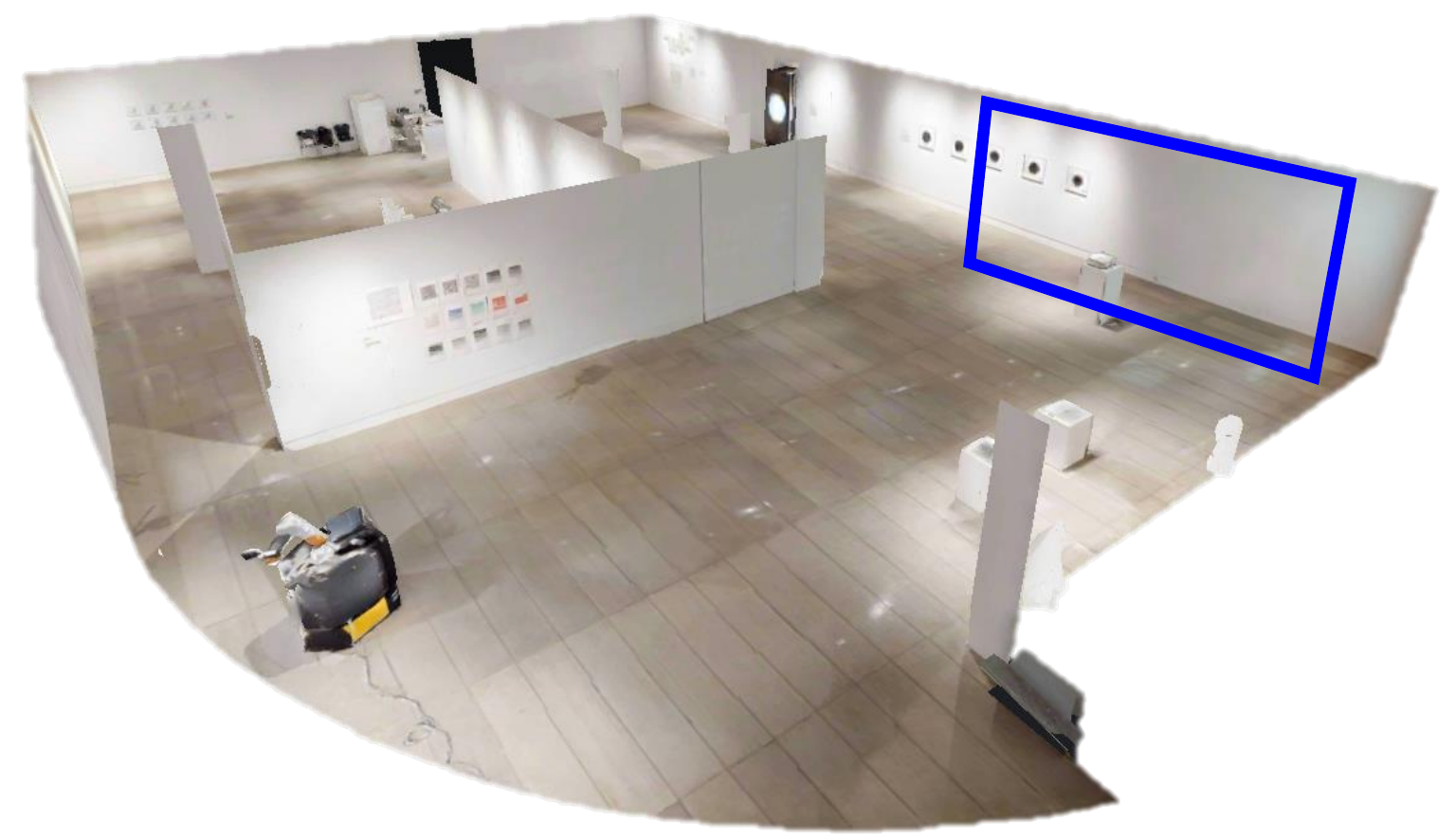}}
\subfigure[Zoomed-in view]
{\label{fig:artp}\includegraphics[width=0.45\columnwidth]{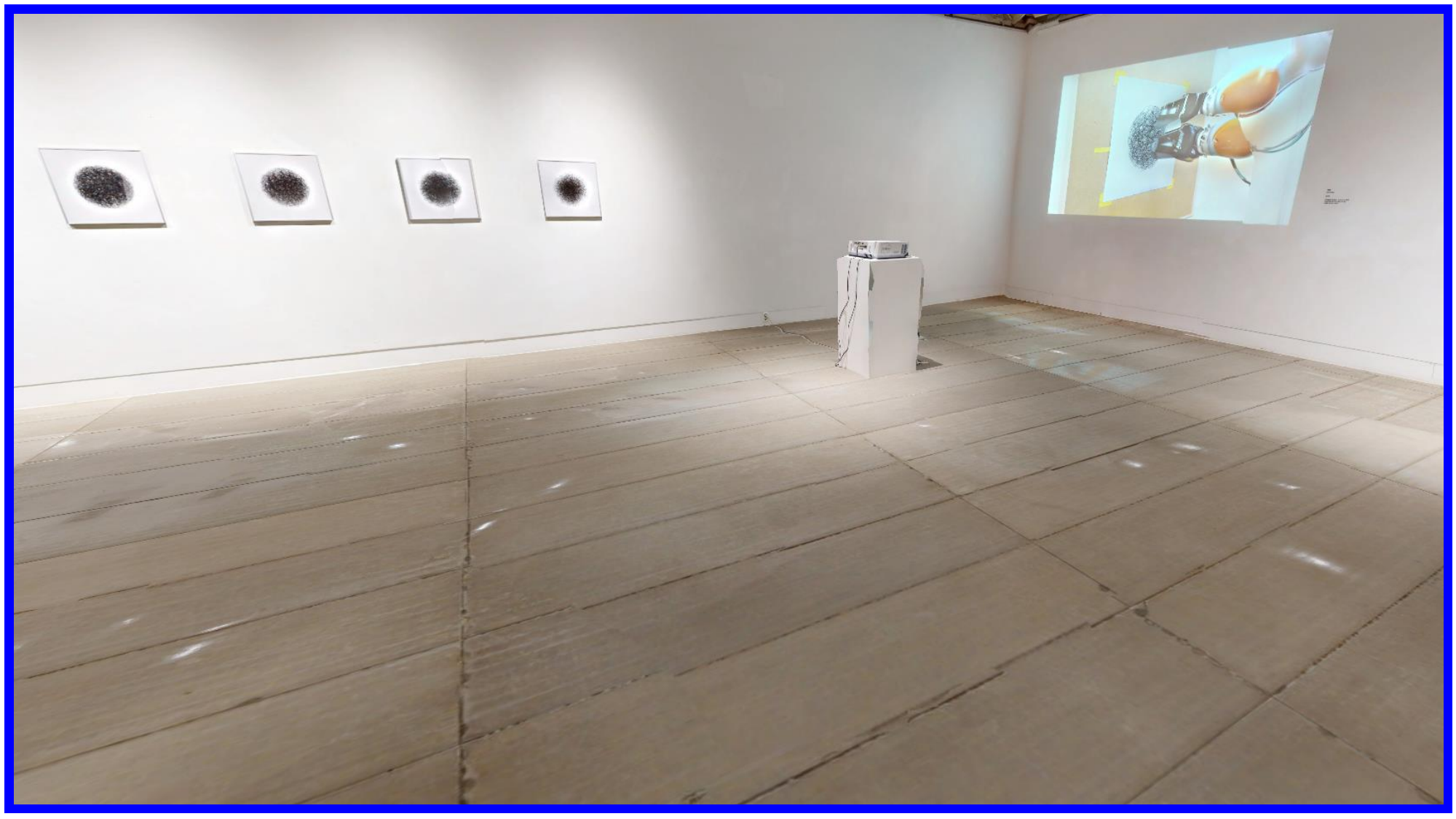}}
\caption{ \textbf{The 3D models reconstructed from various environments.} 
(a), (d), and (e) show the 3D models, while (b) and (c) present zoomed-in views of the colored squares in (a). (f) presents a zoomed-in view of the blue square in (e). }   \label{fig:recon-results}
\end{centering}
\vspace*{-1.0em}
\end{figure}

In a real-world scenario, the robot must rely on estimated SLAM, which is prone to errors during long-term navigation. Additionally, our method efficiently selects viewpoints away from obstacles, thanks to our viewpoint evaluation function (Eq. \ref{eq:viewpoint_eval}), whereas other approaches do not account for such factors when generating trajectories. Note that visiting viewpoints near obstacles can lead to unintended collisions or unnecessary motion behaviors, increasing the likelihood of SLAM errors. This negatively impacts the scanning quality and the final 3D model, as shown in the second row of Fig. \ref{fig:prop-l}. Lastly, Fig. \ref{fig:recon-results} shows three more instances of the 3D models obtained from our system. 
\vspace*{-0.5em}

\section{Conclusion}\label{section:Conclusion}
We introduced a novel autonomous view planning system for panoramic RGB-D cameras. Our visibility-based set-covering algorithm selects the minimum set of viewpoints, which are then used to generate efficient tour plans. The system achieved nearly 99\% coverage while outperforming other state-of-the-art approaches in terms of view planning efficiency. 

In the future, we plan to unify the exploration and scanning phases into a single process, enabling more efficient scanning in unknown environments. Moreover, we aim to extend our system to outdoor robots, developing a more generalized approach for scanning indoor and outdoor environments.

\bibliographystyle{IEEEtran}
\bibliography{references}

\end{document}